\pdfoutput=1

\documentclass[11pt]{article}

\usepackage[preprint]{acl}

\usepackage{times}
\usepackage{latexsym}

\usepackage[T1]{fontenc}

\usepackage[utf8]{inputenc}

\usepackage{microtype}

\usepackage{inconsolata}

\usepackage{graphicx}

\usepackage{bm}
\usepackage{amsmath}
\usepackage{amsfonts}
\usepackage{graphicx}
\usepackage{multirow}
\usepackage{caption}
\usepackage{subcaption}

\DeclareMathOperator*{\diag}{diag}
\DeclareMathOperator*{\LN}{LN}

\title{Fast-and-Frugal Text-Graph Transformers are Effective Link Predictors}

\author{
Andrei C. Coman \\
Idiap Research Institute, EPFL \\
\href{mailto:andrei.coman@idiap.ch}{andrei.coman@idiap.ch} \\
\And
Christos Theodoropoulos \\
KU Leuven \\
\href{mailto:christos.theodoropoulos@kuleuven.be}{christos.theodoropoulos@kuleuven.be}
\AND
Marie-Francine Moens \\
KU Leuven \\
\href{mailto:sien.moens@kuleuven.be}{sien.moens@kuleuven.be}
\And
James Henderson \\
Idiap Research Institute \\
\href{mailto:james.henderson@idiap.ch}{james.henderson@idiap.ch}
}

\begin{document}
\maketitle
\begin{abstract}
We propose Fast-and-Frugal Text-Graph (FnF-TG) Transformers, a Transformer-based framework that unifies textual and structural information for inductive link prediction in text-attributed knowledge graphs. We demonstrate that, by effectively encoding ego-graphs (1-hop neighbourhoods), we can reduce the reliance on resource-intensive textual encoders. This makes the model both fast at training and inference time, as well as frugal in terms of cost. We perform a comprehensive evaluation on three popular datasets and show that FnF-TG can achieve superior performance compared to previous state-of-the-art methods. We also extend inductive learning to a fully inductive setting, where relations don't rely on transductive (fixed) representations, as in previous work, but are a function of their textual description. Additionally, we introduce new variants of existing datasets, specifically designed to test the performance of models on unseen relations at inference time, thus offering a new test-bench for fully inductive link prediction.
\end{abstract}

\section{Introduction}
\label{sec:introduction}

Knowledge graphs (KGs) represent complex information as a structured collection of entities and their relations. They are a fundamental component of various applications, including information extraction \citep{mintz-etal-2009-distant, bosselut-etal-2019-comet, theodoropoulos-etal-2021-imposing} and retrieval \citep{10.1145/2600428.2609628, gupta-etal-2019-care}, question answering \citep{saxena-etal-2022-sequence, yu-etal-2022-kg, coman-etal-2023-strong}, reasoning \citep{zhang-etal-2020-grounded, jiang2022unikgqa, niu-etal-2022-cake}, fact-aware language modelling \citep{logan-etal-2019-baracks, Yang2023GiveUT}, and many others \citep{Fensel2020KnowledgeGM, schneider-etal-2022-decade}. Text-attributed KGs extend KGs by associating each entity and relation with a corresponding textual description, which provide a richer representation of the knowledge encoded in the graph.  In particular, the text associated with an entity may provide a description of its relationships to other entities.
This combination of explicit structural and implicit textual information makes modelling text-attributed KGs particularly challenging.

Initial attempts to model KGs focused on their graph nature, typically addressing a \textit{transductive} setting  \citep{NIPS2013_1cecc7a7, nickel2015review, 8047276}. These models could only make predictions for entities observed during training and only considered the structural information of the KG, ignoring any textual information.

To overcome this limitation, later work focused on using the textual descriptions in KGs to address an \textit{inductive} setting \citep{Xie2016RepresentationLO, shi2018open, wang-etal-2021-kepler}, meaning that predictions can be made even for entities not observed during training, using entity representations computed based on their textual descriptions.

Combining information from textual descriptions and graph structures has proven crucial \citep{Schlichtkrull2017ModelingRD}. An entity's ego-graph, which represents it's 1-hop neighbourhood, provides valuable context that can help disambiguate its role and distinguish it from similar entities. While there has been progress in leveraging ego-graphs, we believe that there is significant room for more effective approaches.

Modelling text-attributed KGs in an inductive setting poses several challenges, particularly when it comes to effectively integrating textual and structural information in embeddings. Transformers \citep{NIPS2017_3f5ee243} have shown remarkable success at modelling unstructured (text) data \citep{devlin-etal-2019-bert, Raffel2019ExploringTL, NEURIPS2020_1457c0d6, Touvron2023LLaMAOA}. While their ability to model structured (graph) data is less evident, the inherent graph processing abilities of a Transformer's self-attention mechanism make it a natural fit for modelling graph structures \citep{henderson-etal-2023-transformers}. We leverage recent advances in using Transformers for graph encoding \citep{mohammadshahi-henderson-2020-graph, mohammadshahi-henderson-2021-recursive, miculicich-henderson-2022-graph, coman-etal-2024-gadepo}, and propose Fast-and-Frugal Text-Graph (FnF-TG) Transformers, which unify textual and structural information in a framework based solely on Transformers.

Another challenge is that text encoders are resource-intensive, especially when the textual descriptions of both entities and their ego-graph neighbours need to be encoded, leading to considerably increased training and inference time \citep{markowitz-etal-2022-statik}. This cost can be reduced by using smaller text encoders, but they can be considerably less effective.
We demonstrate that we can reduce the dependence on large text encoders with a more effectively encoding of ego-graphs, using our FnF-TG Transformers and their more appropriate inductive biases. This makes the overall framework both fast in terms of time, as well as frugal in terms of cost.

A third challenge is that previous models fail to leverage the textual descriptions of relation labels. They still assume a fixed (transductive) inventory of relations, meaning that they cannot handle relations which they did not see during training and thus are not fully inductive. We propose an extension of this method to address the challenge of being fully inductive, by computing a relation embedding from the text describing that relation. This embedding serves as both the relation representation for link prediction, analogous to the transductive case, and also as the relation label which is input to the self-attention mechanism of the FnF-TG's graph encoder component.

We showcase the effectiveness of our proposed model on three popular datasets for inductive link prediction in text-attributed KGs from the experimental setting of \citet{10.1145/3442381.3450141} and \citet{wang-etal-2021-kepler}, namely WN18RR\textsubscript{\textsc{IND}}, FB15k-237\textsubscript{\textsc{IND}}, and Wikidata-5M\textsubscript{\textsc{IND}}.  We show that it improves over the state-of-the-art in all cases.

Additionally, we introduce new variants of existing datasets which are specifically designed to evaluate the performance of models on relations which are unseen until test time, thus offering a new test-bench for fully inductive link prediction.

\paragraph{Contributions:}
\begin{enumerate}
    \vspace{-1.5ex}
    \addtolength{\itemsep}{-1.5ex}
    \item We propose a KG embedding model which leverages the intrinsic graph processing capabilities of Transformers to effectively capture the information in both the KG's textual descriptions and the KG's graph structure.
    \item We demonstrate that Fast-and-Frugal Text-Graph (FnF-TG\footnote{\href{https://github.com/idiap/fnf-tg}{https://github.com/idiap/fnf-tg}}) Transformers achieve superior performance compared to previous state-of-the-art results on three popular
    datasets, even with small and efficient text encoders.
    \item We extend inductive KG learning to a fully inductive setting, where both entity and relation representations are computed as functions of their textual descriptions.
    \item We introduce a new test-bench for fully inductive link prediction by modifying existing datasets to specifically test models' performance on unseen relations.
\end{enumerate}

\section{Related Work}
\label{sec:related_work}

\paragraph{Transductive Link Prediction} In this setting, link prediction aims to identify missing links within a fixed and fully observable graph where all entities and their other connections are known during training. Typically, it involves learning embeddings within a geometric space, as demonstrated by models like RESCAL \citep{10.5555/3104482.3104584}, NTN \citep{NIPS2013_b337e84d}, TransE \citep{NIPS2013_1cecc7a7}, DistMult \citep{yang2014embedding}, ComplEx \citep{trouillon2016complex}, TorusE \citep{10.5555/3504035.3504257}, RotatE \citep{sun2019rotate}, and SimplE \citep{NEURIPS2018_b2ab0019}. Additionally, there are approaches that incorporate convolutional layers such as R-GCN \citep{Schlichtkrull2017ModelingRD}, ConvE \citep{10.5555/3504035.3504256}, HypER \citep{Balazevic2018HypernetworkKG}, and ConvR \cite{jiang-etal-2019-adaptive}. Moreover, recent advances have seen the integration of Transformers in models such as CoKE \citep{wang2019coke} and HittER \citep{chen-etal-2021-hitter}.

\paragraph{Inductive Link Prediction} In this setting, link prediction involves predicting missing links in a dynamic graph where only partial information is available during training. Much work has explored leveraging limited relational knowledge between novel entities and those already present in the training graph \citep{bhowmik2020explainable, Wang2020RelationalMP}. Examples include LAN \citep{10.1609/aaai.v33i01.33017152}, IndTransE \citep{dai-etal-2021-inductively}, OpenWorld \citep{shah2019open}, GraIL \citep{10.5555/3524938.3525814}, NBFNet \citep{Zhu2021NeuralBN}, NodePiece \citep{Galkin2021NodePieceCA}, and BERTRL \citep{Zha2021InductiveRP}. Moreover, approaches such as DKRL \citep{Xie2016RepresentationLO}, Commonsense \citep{malaviya2020commonsense}, KG-BERT \citep{Yao2019KGBERTBF}, KEPLER \citep{wang-etal-2021-kepler}, BLP \citep{10.1145/3442381.3450141}, StAR \citep{10.1145/3442381.3450043}, SimKGC \citep{wang-etal-2022-simkgc}, StATIK \citep{markowitz-etal-2022-statik}, iHT \citep{Chen2023PretrainingTF}, and KnowC \citep{yang-etal-2024-knowledge} use language models to encode entities based on their textual descriptions. Among these, StATIK \citep{markowitz-etal-2022-statik} stands out as it combines both a language model and a graph encoder, specifically employing a Message Passing Neural Network (MPNN) \citep{10.5555/3305381.3305512} to create entity embeddings. This makes StATIK particularly relevant to our approach and we will use it as the state-of-the-art method of reference and compare our proposed method against it to demonstrate its effectiveness.

\paragraph{Transformers and Graphs} Graph Transformers (GTs) represent a significant evolution in graph input methods within the Transformer architecture \citep{henderson-etal-2023-transformers}. Early work such as G2GT \citep{mohammadshahi-henderson-2020-graph, mohammadshahi-henderson-2021-recursive, miculicich-henderson-2022-graph} laid the foundation by incorporating explicit graphs into Transformer's latent attention graph. Later work introduced RoFormer \citep{Su2021RoFormerET}, which uses a rotation matrix to encode absolute positions, and Graphormer \citep{Ying2021DoTR}, which uses node centrality encoding and soft attention biases. Other models, like SSAN \citep{Xu2021EntitySW}, JointGT \citep{ke-etal-2021-jointgt}, TableFormer \citep{yang-etal-2022-tableformer}, and GADePo \citep{coman-etal-2024-gadepo}, have applied GTs to various tasks such as document-level relation extraction, knowledge-to-text generation, table-based question answering, and graph-aware declarative pooling.

We continue to advance graph input methods and show that GTs, when combined with an effective inductive bias in the input and the latent attention graph, achieve superior performance compared to the previous state-of-the-art.

\begin{figure*}
  \centering
  \includegraphics[width=1.0\textwidth]{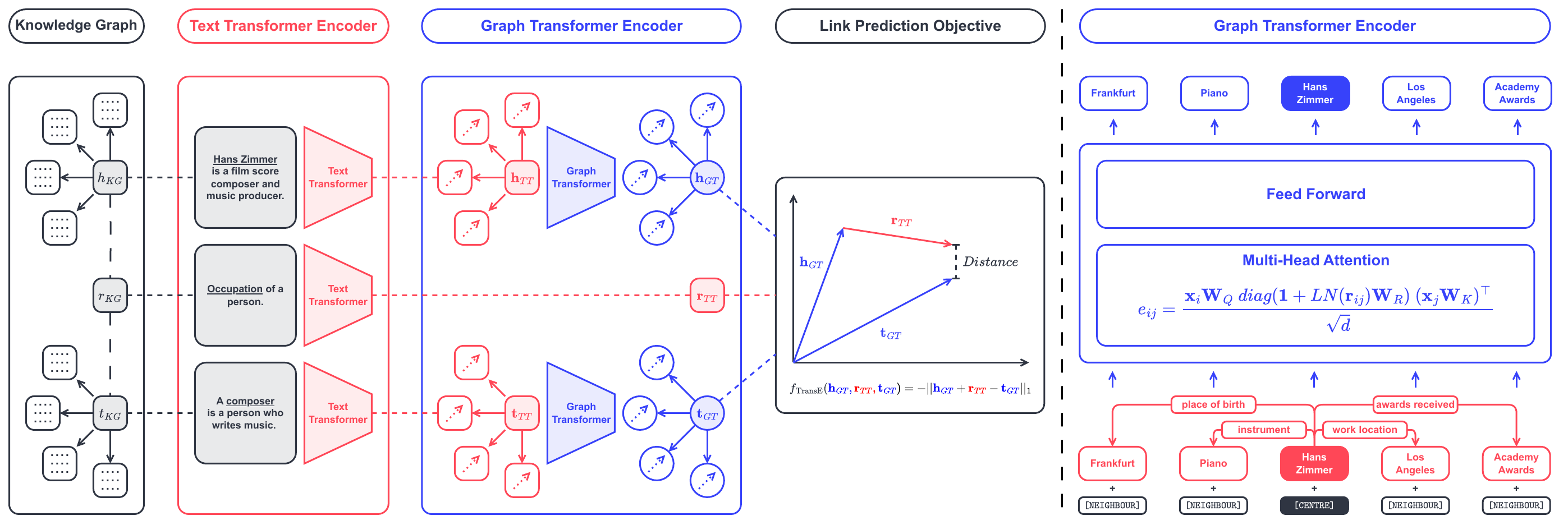}
  \caption{Architecture of the proposed Fast-and-Frugal Text-Graph (FnF-TG) Transformer model.
  }
  \label{fig:fnf}
\end{figure*}

\section{Background}
\label{sec:background}

\subsection{Inductive Representation Learning}
\label{subsec:inductive_representation_learning}

A text-attributed knowledge graph can be defined as $\mathcal{G} = (\mathcal{E}, \mathcal{R}, \mathcal{T}, \mathcal{D})$ where $\mathcal{E}$ represents the set of entities, $\mathcal{R}$ denotes the set of relation labels, $\mathcal{T}$ consists of the set of relation triples $(h,r,t) \in \mathcal{E} \times \mathcal{R} \times \mathcal{E}$, and $\mathcal{D}$ contains the textual descriptions associated with entities and relation labels. In each triple, $h$ and $t$ represent the head and tail entities, respectively, which are connected by a directional relation $r$. Inductive link prediction involves completing missing triples in the graph by leveraging the textual descriptions associated with the entities and relation labels.  If the textual description of an entity mentions the target relation, then this resembles an open relation extraction task \citep{Banko2007OpenIE}, and if not then it resembles a knowledge graph completion task \citep{Lin_Liu_Sun_Liu_Zhu_2015}, with a continuum of difficulties in between.  Our goal is to learn an embedder that maps these descriptions and the partial graph to a representation space where the missing triples can be inferred. Specifically, given a training graph $\mathcal{G}_{train} = (\mathcal{E}_{train}, \mathcal{R}, \mathcal{T}_{train}, \mathcal{D}_{train})$, where $\mathcal{E}_{train} \subset \mathcal{E}$, $\mathcal{D}_{train} \subset \mathcal{D}$ and $\mathcal{T}_{train} \subset \mathcal{T}$ only includes triples involving entities in $\mathcal{E}_{train}$, the goal is to infer the missing triples in $\mathcal{T}\setminus\mathcal{T}_{train}$. During evaluation, for a given query triple $\mathcal{T}_{i} = (h,r,t)$, the model is tasked with performing head or tail prediction on the graph $\mathcal{G}\setminus\mathcal{T}_{i}$. This involves two types of queries: tail prediction, where the query is of the form $(h,r,\hat{e})$ and head prediction, where the query is of the form $(\hat{e}, r, t)$. In both cases, the model must rank all possible candidate entities $\hat{e} \in \hat{\mathcal{E}}$ to identify the correct entity $\hat{e}_{t}$ or $\hat{e}_{h}$ and place it at the top of the ranked list.

\subsection{Structural Objective and Loss Function}
\label{subsec:structural_objective_and_loss_function}

We adopt the margin-based ranking loss from \citet{NIPS2013_1cecc7a7} as our optimisation criterion. We construct two sets of triples: a set of true triples $T$ and a set of negative triples $T'$, where a negative triple consists of a corrupted version of a true triple with either the head or tail entity replaced by a random entity (target excluded) from the training minibatch. We define the structural objective function $f$ using the TransE \citep{NIPS2013_1cecc7a7} model, which represents each triple $(h, r, t)$ as:
\vspace{-1ex}
\begin{equation*}
    f_{\text{TransE}}(h,r,t) = -||\bm{h} + \bm{r} - \bm{t}||_{1}
\vspace{-1ex}
\end{equation*}
where $\bm{h}$, $\bm{r}$, and $\bm{t}$ are the vector representations of the head entity, relation, and tail entity, respectively. The loss function is then defined as:
\vspace{-1ex}
\begin{equation*}
    Loss = \sum_{(t, t') \in (T \times T')} max(0, 1 - f(t) + f(t'))
\vspace{-1ex}
\end{equation*}
where $f(t)$ and $f(t')$ are the scores assigned to the true triple $t$ and the negative triple $t'$, respectively.

\subsection{Evaluation and Metrics}

\paragraph{Evaluation Scenarios} We assess our models in two inductive scenarios, following \citet{NIPS2013_1cecc7a7}. In the first setting, called \textit{dynamic} evaluation, new entities may appear in the head or tail positions, and the candidates set is defined as $\hat{\mathcal{E}} = \mathcal{E}_{train} \cup \mathcal{E}_{eval}$. In the second setting, called \textit{transfer} evaluation, both head and tail entities are new and unseen during training, and the candidates set is defined as $\hat{\mathcal{E}} = \mathcal{E}_{eval}$, where $\mathcal{E}_{eval}$ is disjoint from the training set of entities $\mathcal{E}_{train}$.

\paragraph{Metrics} For each evaluation triple, we create two types of queries: $(h, r, \hat{e})$ for predicting tails and $(\hat{e}, r, t)$ for predicting heads, where $\hat{e} \in \mathcal{E}$ represents all possible candidate entities, as described in Subsection \ref{subsec:inductive_representation_learning}. We rank candidate triples by their scores and evaluate the ranking of the correct triple. We report Mean Reciprocal Rank (MRR) and Hits@k (H@k) with $k \in \{1, 3, 10\}$ averaged across head and tail prediction tasks. We adopt the \textit{filtered} setting as in \citet{NIPS2013_1cecc7a7}, removing valid triples from the set of negative candidate triples when ranking candidate targets.

\subsection{Proposed Architecture}
\label{subsection:proposed_architecture}

Figure \ref{fig:fnf} shows the overall architecture of our proposed model. There are three main components: \textit{Knowledge Graph} (KG), \textit{Text Transformer Encoder} (TT) and \textit{Graph Transformer Encoder} (GT).

\paragraph{Knowledge Graph} The text-attributed KG component contains a set of triples of type $(h_{KG}, r_{KG}, t_{KG})$, along with their corresponding textual descriptions. For each head $h_{KG}$ and tail $t_{KG}$, we also extract their ego-graphs (1-hop neighbourhood), denoted as $E(h_{KG})$ and $E(t_{KG})$, respectively. Then we encode each of these nodes with the text encoder, discussed below. Specifically, for each centre entity, we encode its textual description, as well as encoding the textual descriptions of its neighbouring entities and the relations that connect the centre entity to its neighbours.

\paragraph{Text Transformer Encoder} The textual descriptions from the KG module are passed to the \textit{Text Transformer Encoder} (TT), which produces vector representations $\bm{x}_{TT}$ for each entity and relation textual description $x_{KG}$ in the ego-graph. More formally, we apply the following function:
\vspace{-1ex}
\begin{equation*}
    \bm{x}_{TT} = \sigma(\text{BERT}\textsubscript{\textsc{size}}(x_{KG})_{\texttt{[CLS]}}\bm{W}_{0})\bm{W_{1}}
\vspace{-1ex}
\end{equation*}
where $\bm{W}_{0}, \bm{W}_{1} \in \mathbb{R}^{d \times d}$ are two linear projection matrices and $\sigma$ is the SiLU \citep{Elfwing2017SigmoidWeightedLU} activation function. We employ BERT \cite{devlin-etal-2019-bert} as our encoder and use the \texttt{[CLS]} vector representation output by the encoder as the embedding. BERT\textsubscript{\textsc{size}} indicates that we employ different sizes of this model released by \citet{Turc2019WellReadSL}.

When encoding candidate entities $\hat{e} \in \hat{\mathcal{E}}$, we simply pass the text associated with the entity through the TT component.  The same is true for any neighbouring entities required by the graph encoder (discussed below).  In contrast, when encoding queries $(h, r, \cdot)$ and $(\cdot, r, t)$ we condition $h$ and $t$ on the relation type $r$. Similarly to StAR \citep{10.1145/3442381.3450043} and StATIK \citep{markowitz-etal-2022-statik}, for tail prediction queries $(h, r, \cdot)$ we concatenate the text associated with $h_{KG}$ and $r_{KG}$, resulting in $[h||r]_{KG}$. For head prediction queries $(\cdot, r, t)$, we create an inverse version of the relation text by prepending its textual description with the text \texttt{"inverse of"}, denoted as $r^{-1}_{KG}$. We then concatenate the text associated with $t_{KG}$ and $r^{-1}_{KG}$, resulting in $[t||r^{-1}]_{KG}$.

\paragraph{Graph Transformer Encoder} The outputs of the TT component, together with the ego-graphs $E(\bm{h}_{TT})$ and $E(\bm{t}_{TT})$ are then input to the \textit{Graph Transformer Encoder} (GT). The input embedding for each node of the graph is the vector output by the TT encoder for that entity, as described above. In addition, we add learnable segment embeddings to each node input, indicated as $s_{\texttt{[CENTRE]}}$ and $s_{\texttt{[NEIGHBOUR]}}$, to disambiguate between the centre and neighbour nodes in the ego-graph. These embeddings indicate to the model which input nodes will be used subsequently as the embedding representation of the ego-graph.

To encode the graph relations, we follow \citet{mohammadshahi-henderson-2020-graph, mohammadshahi-henderson-2021-recursive, miculicich-henderson-2022-graph, coman-etal-2024-gadepo} in leveraging the intrinsic graph processing capabilities of the Transformer model by incorporating graph relations as relation embeddings input to the self-attention function. But unlike in that previous work, our relation embeddings are computed from the text associated with the relation, rather than coming from a fixed set of relations. For every pair of input nodes $ij$, the pre-softmax attention score $e_{ij} \in \mathbb{R}$ is computed from both the respective node embeddings $\bm{x}_{i}, \bm{x}_{j} \in \mathbb{R}^{d}$, and the embedding of the relation $r_{ij}$ between the $i$-th and $j$-th nodes, as:
\vspace{-1ex}
\begin{equation*}
e_{ij} = \frac{ \bm{x}_i\bm{W}_Q ~ \diag(\bm{1} + \LN(\bm{r}_{ij})\bm{W}_R) ~ (\bm{x}_j\bm{W}_K)^\top }{\sqrt{d}}
\vspace{-1ex}
\end{equation*}
where $\bm{W}_Q, \bm{W}_K \in \mathbb{R}^{d \times d}$ represent the query and key matrices, respectively, $\bm{r}_{ij}$ represents the relation embedding output by the TT module when it encodes the text associated with the relation between the $i$-th and $j$-th nodes, and $\bm{W}_R \in \mathbb{R}^{d \times d}$ is the relation matrix. Thus, $\LN(\bm{r}_{ij})\bm{W}_R$ is the embedding of the relation between $i$ and $j$, where $\LN$ stands for the $LayerNorm$ operation.  Finally, $\diag(\bm{1}+\ldots)$ maps this vector into a diagonal matrix plus the identity matrix.

When encoding candidate entities and queries in the GT, it is crucial to ensure that no information regarding the target triple $(h, r, t)$ leaks into the $E(\hat{e})$, $E(\bm{h}_{TT})$, or $E(\bm{t}_{TT})$ ego-graphs. This precaution prevents the model from learning trivial solutions or biases from leaked information.

\subsection{Datasets and Setting}

\citet{10.1145/3442381.3450141} introduced the WN18RR\textsubscript{\textsc{IND}} and FB15k-237\textsubscript{\textsc{IND}} inductive variants of the well-known WN18RR\textsubscript{\textsc{TRA}} \citep{10.5555/3504035.3504256} and FB15k-237\textsubscript{\textsc{TRA}} \citep{toutanova-chen-2015-observed} transductive KGs. The inductive setting simulates a \textit{dynamic} scenario where new entities and triples are dynamically added to the graph.
The training graph is constructed as $\mathcal{G}_{train} = \{(h,r,t) \in \mathcal{T}_{train} : h,t \in \mathcal{E}_{train}\}$. The validation and test graphs are constructed by incrementally adding entities and triples, such that $\mathcal{G}_{val} = \{(h,r,t) \in \mathcal{T}_{val} : h,t \in \mathcal{E}_{train} \cup \mathcal{E}_{val}\}$ and $\mathcal{G}_{test} = \{(h,r,t) \in \mathcal{T}_{test} : h,t \in \mathcal{E}_{train} \cup \mathcal{E}_{val} \cup \mathcal{E}_{test}\}$.

In contrast, the Wikidata5M\textsubscript{\textsc{IND}} KGs curated by \citet{wang-etal-2021-kepler} provide a \textit{transfer} learning scenario in which the evaluation graphs are constructed such that the validation and test entity and triple sets, $\mathcal{E}_{val}$ and $\mathcal{E}_{test}$, and $\mathcal{T}_{val}$ and $\mathcal{T}_{test}$ are disjoint from the training entity and triple set $\mathcal{E}_{train}$ and $\mathcal{T}_{train}$. The validation and test graphs are constructed as $\mathcal{G}_{val} = \{(h,r,t) \in \mathcal{T}_{val} : h,t \in \mathcal{E}_{val}\}$ and $\mathcal{G}_{test} = \{(h,r,t) \in \mathcal{T}_{test} : h,t \in \mathcal{E}_{test}\}$. 
Because the graphs $\mathcal{G}_{val}$ and $\mathcal{G}_{test}$ do not include the entities from $\mathcal{G}_{train}$, they are much smaller graphs (see Appendix Table \ref{tab:datasets}), which poses challenges for generalisation with graph-aware models, as will be discussed further below. We evaluate our model's ability to generalise to entirely new entities and triples in this setting.

We conduct our experiments in the above-mentioned settings of \citet{10.1145/3442381.3450141} and \citet{wang-etal-2021-kepler}, where textual information extraction is an integral part. Our method is directly comparable to DKRL \citep{Xie2016RepresentationLO}, BLP \citep{10.1145/3442381.3450141}, KEPLER \citep{wang-etal-2021-kepler}, StAR \citep{10.1145/3442381.3450043}, and the state-of-the-art method, StATIK \citep{markowitz-etal-2022-statik} which employ the same textual encoder, structural objective, and loss function. Similar to StATIK, our work aims to jointly model the text and the structure of knowledge graphs, including extracting information about KG links from the text. This sets us apart from the setting of \citet{10.5555/3524938.3525814}, which uses different KGs splits and is employed in GraIL \citep{10.5555/3524938.3525814}, NBFNet \citep{Zhu2021NeuralBN}, and NodePiece \citep{Galkin2021NodePieceCA}, that solely focus on using the structure of the graph without incorporating any textual information extraction component.

\begin{table*}[tb]
  \centering
  \scalebox{0.8}{
      \begin{tabular}{l|cccc|cccc}
        \hline
        \multicolumn{1}{}{} & \multicolumn{4}{c}{\textbf{WN18RR\textsubscript{\textsc{IND}}}} & \multicolumn{4}{c}{\textbf{FB15k-237\textsubscript{\textsc{IND}}}} \\
        \textbf{Model} & \textbf{MRR} & \textbf{H@1} & \textbf{H@3} & \textbf{H@10} & \textbf{MRR} & \textbf{H@1} & \textbf{H@3} & \textbf{H@10} \\
        \hline
        \multicolumn{1}{}{} & \multicolumn{8}{c}{\textit{Text-Only Models}} \\
        $\text{DKRL}^{\ast}_{\text{BERT}\textsubscript{\textsc{base}}}$ & $0.139$ & $0.048$ & $0.169$ & $0.320$ & $0.144$ & $0.084$ & $0.151$ & $ 0.263$ \\
        $\text{BOW}^{\ast}_{\text{BERT}\textsubscript{\textsc{base}}}$ & $0.180$ & $0.045$ & $0.244$ & $0.450$ & $0.173$ & $0.103$ & $0.184$ & $0.316$ \\
        $\text{BLP}^{\ast}_{\text{BERT}\textsubscript{\textsc{base}}}$ & $0.285$ & $0.135$ & $0.361$ & $0.580$ & $0.195$ & $0.113$ & $0.213$ & $0.363$ \\
        \hline
        $\text{FnF-T}_{\text{BERT}\textsubscript{\textsc{base}}}$ & $\textbf{0.373}$ & $\textbf{0.238}$ & $\textbf{0.442}$ & $\textbf{0.647}$ & $\textbf{0.266}$ & $\textbf{0.174}$ & $\textbf{0.297}$ & $\textbf{0.453}$ \\
        $\text{FnF-T}_{\text{BERT}\textsubscript{\textsc{medium}}}$ & $0.342$ & $0.213$ & $0.405$ & $0.603$ & $0.253$ & $0.164$ & $0.281$ & $0.431$ \\
        $\text{FnF-T}_{\text{BERT}\textsubscript{\textsc{small}}}$ & $0.320$ & $0.197$ & $0.379$ & $0.572$ & $0.239$ & $0.152$ & $0.265$ & $0.411$ \\
        $\text{FnF-T}_{\text{BERT}\textsubscript{\textsc{mini}}}$ & $0.268$ & $0.156$ & $0.318$ & $0.498$ & $0.204$ & $0.128$ & $0.223$ & $0.354$ \\
        $\text{FnF-T}_{\text{BERT}\textsubscript{\textsc{tiny}}}$ & $0.193$ & $0.098$ & $0.230$ & $0.385$ & $0.164$ & $0.100$ & $0.176$ & $0.289$ \\
        \hline
        \hline
        \multicolumn{1}{}{} & \multicolumn{8}{c}{\textit{Structure-Informed Models}} \\
        $\text{StAR}^{\star}_{\text{BERT}\textsubscript{\textsc{base}}}$ & $0.321$ & $0.192$ & $0.381$ & $0.576$ & $0.163$ & $0.092$ & $0.176$ & $0.309$ \\
        $\text{StATIK}^{\star}_{\text{BERT}\textsubscript{\textsc{base}}}$ & $0.516$ & $0.425$ & $0.558$ & $0.690$ & $0.224$ & $0.143$ & $0.248$ & $0.381$ \\
        \hline
        $\text{FnF-TG}_{\text{BERT}\textsubscript{\textsc{base}}}$ & $0.732$ & $0.652$ & $0.785$ & $\textbf{0.875}$ & $0.316$ & $0.214$ & $0.350$ & $\textbf{0.524}$ \\
        $\text{FnF-TG}_{\text{BERT}\textsubscript{\textsc{medium}}}$ & $\textbf{0.737}$ & $\textbf{0.661}$ & $\textbf{0.789}$ & $0.873$ & $0.314$ & $0.214$ & $0.353$ & $0.515$ \\
        $\text{FnF-TG}_{\text{BERT}\textsubscript{\textsc{small}}}$ & $0.727$ & $0.648$ & $0.781$ & $0.867$ & $\textbf{0.316}$ & $\textbf{0.216}$ & $\textbf{0.354}$ & $0.518$ \\
        $\text{FnF-TG}_{\text{BERT}\textsubscript{\textsc{mini}}}$ & $0.713$ & $0.632$ & $0.768$ & $0.857$ & $0.302$ & $0.204$ & $0.337$ & $0.502$ \\
        $\text{FnF-TG}_{\text{BERT}\textsubscript{\textsc{tiny}}}$ & $0.638$ & $0.543$ & $0.700$ & $0.808$ & $0.288$ & $0.195$ & $0.318$ & $0.475$ \\
        \hline
      \end{tabular}
  }
  \caption{
    WN18RR\textsubscript{\textsc{IND}} and FB15k-237\textsubscript{\textsc{IND}} test set results. $^{\ast}$\citet{10.1145/3442381.3450141}; $^{\star}$\citet{markowitz-etal-2022-statik}.
  }
  \label{tab:inductive}
\end{table*}

\subsection{Controlled Experimental Setup}
\label{subsec:controlled_experimental_setup}

When comparing the performance of different models on link prediction tasks, it is crucial to establish a fair and consistent baseline. Our experiments in Table \ref{tab:controlled_experimental_setup} highlight the importance of carefully setting this baseline, as various factors can greatly influence the results.

\begin{table}[ht]
  \centering
  \scalebox{0.8}{
      \begin{tabular}{l|c|c}
        \hline
        \multicolumn{1}{}{} & \multicolumn{2}{c}{\textbf{MRR}} \\
        \textbf{Model} & \textbf{WN18RR\textsubscript{\textsc{IND}}} & \textbf{FB15k-237\textsubscript{\textsc{IND}}} \\
        \hline
        $\text{BLP}_{\text{BERT}\textsubscript{\textsc{base}}}$ & $0.285$ & $0.195$ \\
        \hline
        $\text{BLP}^{\bullet}_{\text{BERT}\textsubscript{\textsc{base}}}$ & $0.280$ & $0.205$ \\
        $+$ \small inductive relations & $0.281$ & $0.219$ \\
        $+$ \small negatives batch tying & $0.300$ & $0.221$ \\
        $+$ \small bigger embedding size & $0.339$ & $0.254$ \\
        $+$ \small bigger batch size & $0.366$ & $0.260$ \\
        $+$ \small better sampling method& $0.373$ & $0.266$ \\
        \hline
        $\text{FnF-T}_{\text{BERT}\textsubscript{\textsc{base}}}$ (ours) & $\textbf{0.373}$ & $\textbf{0.266}$ \\
        \hline
      \end{tabular}
  }
  \caption{
    WN18RR\textsubscript{\textsc{IND}} and FB15k-237\textsubscript{\textsc{IND}} test set results with cumulative additions over the baseline model $\text{BLP}_{\text{BERT}\textsubscript{\textsc{base}}}$ \citep{10.1145/3442381.3450141} that lead to our improved baseline model $\text{FnF-T}_{\text{BERT}\textsubscript{\textsc{base}}}$. $\text{BLP}^{\bullet}_{\text{BERT}\textsubscript{\textsc{base}}}$ indicates our reimplementation of $\text{BLP}_{\text{BERT}\textsubscript{\textsc{base}}}$.
  }
  \label{tab:controlled_experimental_setup}
\end{table}


\begin{table}[tb]
  \centering
  \scalebox{0.8}{
      \begin{tabular}{l|cccc}
        \hline
        \textbf{Model} & \textbf{MRR} & \textbf{H@1} & \textbf{H@3} & \textbf{H@10} \\
        \hline
        \multicolumn{1}{}{} & \multicolumn{4}{c}{\textit{Text-Only Models}} \\
        $\text{KEPLER}^{\diamond}_{\text{BERT}\textsubscript{\textsc{base}}}$ & $0.402$ & $0.222$ & $0.514$ & $0.730$ \\
        $\text{BLP}^{\ast}_{\text{BERT}\textsubscript{\textsc{base}}}$ & $0.478$ & $0.241$ & $0.660$ & $0.871$ \\
        \hline
        $\text{FnF-T}_{\text{BERT}\textsubscript{\textsc{base}}}$ & $\textbf{0.597}$ & $\textbf{0.427}$ & $\textbf{0.722}$ & $\textbf{0.896}$ \\
        $\text{FnF-T}_{\text{BERT}\textsubscript{\textsc{medium}}}$ & $0.588$ & $0.418$ & $0.712$ & $0.890$ \\
        $\text{FnF-T}_{\text{BERT}\textsubscript{\textsc{small}}}$ & $0.588$ & $0.417$ & $0.714$ & $0.889$ \\
        $\text{FnF-T}_{\text{BERT}\textsubscript{\textsc{mini}}}$ & $0.562$ & $0.391$ & $0.683$ & $0.870$ \\
        $\text{FnF-T}_{\text{BERT}\textsubscript{\textsc{tiny}}}$ & $0.526$ & $0.348$ & $0.649$ & $0.849$ \\
        \hline
        \hline
        \multicolumn{1}{}{} & \multicolumn{4}{c}{\textit{Structure-Informed Models}} \\
        $\text{StATIK}^{\star}_{\text{BERT}\textsubscript{\textsc{base}}}$ & $0.770$ & $\textbf{0.765}$ & $0.771$ & $0.779$ \\
        \hline
        $\text{FnF-TG}_{\text{BERT}\textsubscript{\textsc{base}}}$ & $\textbf{0.799}$ & $0.741$ & $\textbf{0.833}$ & $\textbf{0.911}$ \\
        $\text{FnF-TG}_{\text{BERT}\textsubscript{\textsc{medium}}}$ & $0.785$ & $0.727$ & $0.817$ & $0.900$ \\
        $\text{FnF-TG}_{\text{BERT}\textsubscript{\textsc{small}}}$ & $0.781$ & $0.721$ & $0.816$ & $0.898$ \\
        $\text{FnF-TG}_{\text{BERT}\textsubscript{\textsc{mini}}}$ & $0.779$ & $0.719$ & $0.814$ & $0.894$ \\
        $\text{FnF-TG}_{\text{BERT}\textsubscript{\textsc{tiny}}}$ & $0.761$ & $0.697$ & $0.799$ & $0.883$ \\
        \hline
      \end{tabular}
  }
  \caption{
    Wikidata5M\textsubscript{\textsc{IND}} test set results. $^{\diamond}$\citet{wang-etal-2021-kepler}; $^{\ast}$\citet{10.1145/3442381.3450141}; $^{\star}$\citet{markowitz-etal-2022-statik}.
  }
  \label{tab:wikidata}
\end{table}

Specifically, we demonstrate that the computational budget, which determines training hyperparameters, can have a substantial impact on model performance. Starting with the baseline model $\text{BLP}_{\text{BERT}\textsubscript{\textsc{base}}}$ \citep{10.1145/3442381.3450141}, we introduce improvements such as using inductive relations, increasing the number of negative triples to match the batch size (negatives batch tying), increasing the embedding dimension from 128 to 768, doubling the batch size from 64 to 128, and modifying the negative sampling strategy to two-sided reflexive, where both head and tail entities are considered as potential negatives. These cumulative improvements lead to the development of a new text-only model baseline, $\text{FnF-T}_{\text{BERT}\textsubscript{\textsc{base}}}$, which shows substantial improvements on both the WN18RR\textsubscript{\textsc{IND}} and FB15k-237\textsubscript{\textsc{IND}} datasets.

To ensure a fair comparison, we fixed our computational budget to a constant in this paper, using a consumer-grade GPU (NVIDIA RTX3090 24GB). This allows for a consistent and reproducible experimental setup, enabling a more accurate assessment of performance. For more details, see Appendix~\ref{sec:appendix}.

\subsection{Inductive Link Prediction Results}

As shown in the top half of Table~\ref{tab:inductive}, for both the WN18RR\textsubscript{\textsc{IND}} and the FB15k-237\textsubscript{\textsc{IND}} datasets, our inductive relation embeddings and the enhanced controlled experimental setup result in improved text-only models. These models rely heavily on having powerful text encoders, as shown by the degradation in performance when using smaller versions of BERT as the text encoder. 

The addition of our graph encoder to the model (bottom half of Table~\ref{tab:inductive}) leads to a substantial increase in link prediction accuracy over the text-only model. We also see that our TG (text-graph) encoder results in substantially better accuracy than the previous state-of-the-art model, StATIK. Interestingly, this more effective use of graph context also has a big impact on the model's dependence on powerful text encoders. Reducing the size of the text encoder (${\text{BERT}\textsubscript{\textsc{base}}} > {\text{BERT}\textsubscript{\textsc{medium}}} > {\text{BERT}\textsubscript{\textsc{small}}} > {\text{BERT}\textsubscript{\textsc{mini}}} > {\text{BERT}\textsubscript{\textsc{tiny}}}$) does result in some degradation of accuracy, but the differences are much smaller than in the text-only case.  Even with a ${\text{BERT}\textsubscript{\textsc{tiny}}}$ text encoder, the graph-aware model performs better than the text-only model with a ${\text{BERT}\textsubscript{\textsc{base}}}$ encoder. This shows that the inductive bias of explicit graph relations can be an effective alternative to extracting the same information from text with a powerful text encoder.

This pattern of results is repeated in the transfer case, shown in
Table~\ref{tab:wikidata}.  Here, the training set is much larger, but the graph in the test set is relatively small with each entity having fewer neighbours (see Appendix Table \ref{tab:datasets}). This reduces the advantage gained from adding an effective graph encoder and the margin of our models' improvement over the text-only models, and over the previous state-of-the-art model, StATIK.\footnote{StATIK has a surprisingly high H@1 score, almost identical to its H@3, H@10 and MRR scores. It is not clear why this is the case. Regardless, our model's MRR, H@3, and H@10 scores are better than StATIK. MRR is the primary evaluation measure since it summarises the entire ranking.} But we still see the same pattern where the size of the text encoder has less effect on accuracy for the graph-aware model.

\subsection{Ablation Study}
\label{subsec:ablation_study}

Table~\ref{tab:ablation} presents results from our ablation studies, showing the impact of removing various design features from our graph-aware model on its accuracy.

\begin{table}[!h]
  \centering
  \scalebox{0.75}{
      \begin{tabular}{l|c|c}
        \hline
        \multicolumn{1}{}{} & \multicolumn{2}{c}{\textbf{MRR}} \\
        \textbf{Model} & \textbf{WN18RR\textsubscript{\textsc{IND}}} & \textbf{FB15k-237\textsubscript{\textsc{IND}}} \\
        \hline
        $\text{FnF-TG}_{\text{BERT}\textsubscript{\textsc{medium | small}}}$ & $0.737$ & $0.316$ \\
        $-$ $r_{ij}$ & $0.733$ & $0.306$ \\
        $-$ $s_{\texttt{[CENTRE]}}$, $s_{\texttt{[NEIGHBOUR]}}$ & $0.677$ & $0.298$ \\
        $-$ $E(\bm{h}_{TT})$, $E(\bm{t}_{TT})$ & $0.480$ & $0.251$ \\
        $-$ $[h||r]_{KG}$, $[t||r^{-1}]_{KG}$ & $0.342$ & $0.239$ \\
        \hline
      \end{tabular}
  }
  \caption{
    Ablation studies on the WN18RR\textsubscript{\textsc{IND}} and FB15k-237\textsubscript{\textsc{IND}} test sets using the top FnF-TG model. Each row indicates the performance after cumulatively removing a specific feature.
  }
  \label{tab:ablation}
\end{table}

\noindent Removing the $r_{ij}$ relation embeddings in the pre-softmax attention score leads to a decline in model performance, with a more substantial drop observed on the FB15K-237\textsubscript{\textsc{IND}} dataset compared to the WN18RR\textsubscript{\textsc{IND}} dataset.
Note that with this modification the model still knows that there is some relation to the neighbours, but does not know its label. Removing the learnable segment embeddings $s_{\texttt{[CENTRE]}}$ and $s_{\texttt{[NEIGHBOUR]}}$ then removes this unlabelled graph structure, which considerably impacts the model's performance. 
Eliminating the ego-graph neighbours 
altogether results in an even more substantial performance drop. Despite this, the model remains competitive as a text-only model compared to the BLP baseline, owing to its ability to leverage relation conditioning features to represent candidate relations.
Finally, removing the relation conditioning $[h||r]_{KG}$ and $[t||r^{-1}]_{KG}$, results in a further notable decrease in performance. Without relation conditioning, the model loses its ability to anticipate the query relation, severely impacting its accuracy.

\subsection{Efficient Text Encoders}

\begin{figure}[!h]
  \centering
  \captionsetup[subfigure]{labelformat=empty}
  \begin{subfigure}{\columnwidth}
    \centering
    \includegraphics[width=0.88\columnwidth]{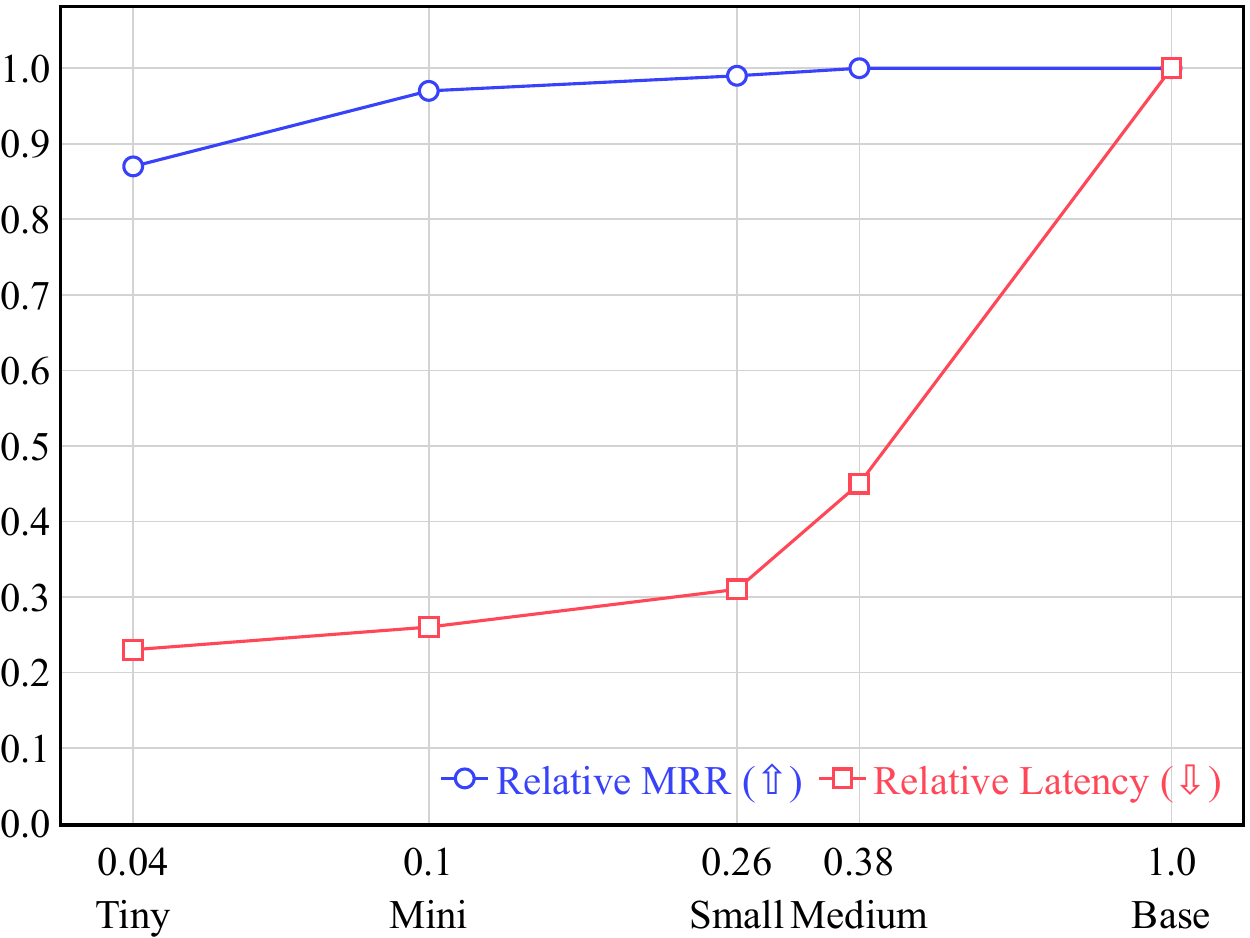}
    \caption{WN18RR\textsubscript{\textsc{IND}}}
  \end{subfigure}%
  \\
  \vspace{0.5em}
  \begin{subfigure}{\columnwidth}
    \centering
    \includegraphics[width=0.88\columnwidth]{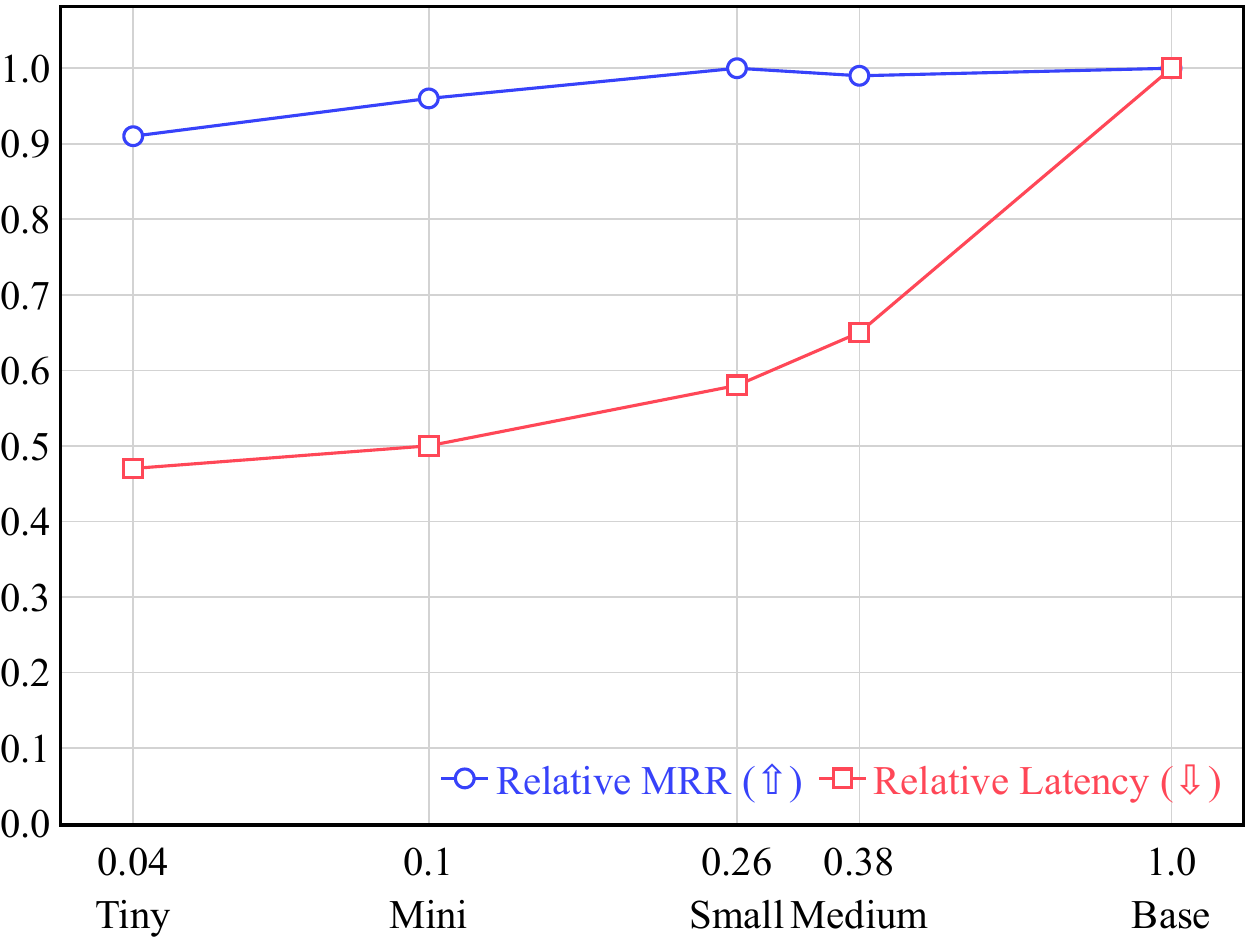}
    \caption{FB15k-237\textsubscript{\textsc{IND}}}
  \end{subfigure}%
  \caption{Accuracy and training time plotted as a function of text encoder size, relative to the largest text encoder with the highest accuracy, shown as $(1.0,1.0)$.
  \vspace{-3ex}
  }
  \label{fig:latency}
\end{figure}

Being able to reduce the size of the text encoder with minimal degradation in accuracy is important because the text encoder is a substantial part of the training cost. In Figure~\ref{fig:latency} we plot the relative reduction in accuracy against the relative reduction in training time as we reduce the size of the text encoder, for the WN18RR\textsubscript{\textsc{IND}} and the FB15k-237\textsubscript{\textsc{IND}} datasets. We see that reducing the encoder size by a factor of four reduces the training time by a factor of three for WN18RR\textsubscript{\textsc{IND}} (and nearly two for FB15k-237\textsubscript{\textsc{IND}}) with very little reduction in accuracy.

\begin{table*}[htb]
  \centering
  \scalebox{0.75}{
      \begin{tabular}{c|c|cccc|cccc|cccc}
        \hline
        \multicolumn{2}{}{} & \multicolumn{4}{c}{\textbf{WN18RR}} & \multicolumn{4}{c}{\textbf{FB15k-237}} & \multicolumn{4}{c}{\textbf{Wikidata5M}} \\
        \textbf{Training} & \textbf{Evaluation} & \textbf{MRR} & \textbf{H@1} & \textbf{H@3} & \textbf{H@10} & \textbf{MRR} & \textbf{H@1} & \textbf{H@3} & \textbf{H@10} & \textbf{MRR} & \textbf{H@1} & \textbf{H@3} & \textbf{H@10} \\
        \hline
        \multicolumn{14}{c}{\multirow{2}{*}{$\text{FnF-T}_{\text{BERT}\textsubscript{\textsc{tiny}}}$}} \\
        \multicolumn{14}{}{} \\
        \hline
        $\textsc{IND}$ & $\textsc{IND}$ & $0.193$ & $0.098$ & $0.230$ & $0.385$ & $0.164$ & $0.100$ & $0.176$ & $0.289$ & $0.526$ & $0.348$ & $0.649$ & $0.849$ \\
        $\textsc{FIR}$ & $\textsc{IND}$ & $0.169$ & $0.080$ & $0.198$ & $0.346$ & $0.143$ & $0.087$ & $0.153$ & $0.253$ & $0.478$ & $0.306$ & $0.591$ & $0.793$ \\
        \hline
        $\textsc{IND}$ & $\textsc{IND}\setminus\textsc{FIR}$ & $0.307$ & $0.210$ & $0.353$ & $0.495$ & $0.171$ & $0.102$ & $0.185$ & $0.308$ & $0.594$ & $0.451$ & $0.697$ & $0.849$ \\
        $\textsc{FIR}$ & $\textsc{IND}\setminus\textsc{FIR}$ & $0.064$ & $0.012$ & $0.082$ & $0.159$ & $0.024$ & $0.007$ & $0.026$ & $0.052$ & $0.219$ & $0.054$ & $0.334$ & $0.499$ \\
        \hline
        \multicolumn{14}{c}{\multirow{2}{*}{$\text{FnF-TG}_{\text{BERT}\textsubscript{\textsc{tiny}}}$}} \\
        \multicolumn{14}{}{} \\
        \hline
        $\textsc{IND}$ & $\textsc{IND}$ & $0.638$ & $0.543$ & $0.700$ & $0.808$ & $0.288$ & $0.195$ & $0.318$ & $0.475$ & $0.761$ & $0.697$ & $0.799$ & $0.883$ \\
        $\textsc{FIR}$ & $\textsc{IND}$ & $0.573$ & $0.480$ & $0.629$ & $0.738$ & $0.249$ & $0.165$ & $0.274$ & $0.418$ & $0.711$ & $0.644$ & $0.749$ & $0.837$ \\
        \hline
        $\textsc{IND}$ & $\textsc{IND}\setminus\textsc{FIR}$ & $0.585$ & $0.483$ & $0.652$ & $0.769$ & $0.282$ & $0.176$ & $0.311$ & $0.514$ & $0.726$ & $0.646$ & $0.781$ & $0.867$ \\
        $\textsc{FIR}$ & $\textsc{IND}\setminus\textsc{FIR}$ & $0.108$ & $0.028$ & $0.147$ & $0.242$ & $0.050$ & $0.023$ & $0.051$ & $0.099$ & $0.401$ & $0.287$ & $0.480$ & $0.589$ \\
        \hline
        \multicolumn{14}{c}{\multirow{2}{*}{Random baseline}} \\
        \multicolumn{14}{}{} \\
        \hline
        $-$ & $-$ & $0.0003$ & $-$ & $-$ & $-$ & $0.0007$ & $-$ & $-$ & $-$ & $0.0013$ & $-$ & $-$ & $-$ \\
        \hline
      \end{tabular}
  }
  \caption{
    Fully inductive link prediction results.
  }
  \label{tab:fully_inductive}
\end{table*}

\subsection{Fully Inductive Link Prediction Results}

The experimental setting of \citet{10.1145/3442381.3450141} and \citet{wang-etal-2021-kepler} do not support evaluation on unseen relations. One distinctive advantage of our model is that it is not restricted to a fixed set of relation labels learned during training. 
Although we do show that conditioning on relation texts improves accuracy even on seen relations (see Table \ref{tab:controlled_experimental_setup}), it is important to evaluate our model in a fully inductive setting, where relations are also unseen, in addition to entities.

To this end, we propose a new experimental setting for a \underline{f}ully \underline{i}nductive \underline{r}elations (FIR) evaluation, by converting the WN18RR\textsubscript{\textsc{IND}}, FB15k-237\textsubscript{\textsc{IND}}, and Wikidata-5M\textsubscript{\textsc{IND}} evaluations to their respective FIR versions. More specifically, we focus on the long tail of relations and remove the least frequent relation labels until 10\% of edges have been removed from the training graph $\mathcal{G}_{train}$. We then train a new set of models on this new version so they have not seen the removed relation labels, and evaluate them on both the full set of test relations ($\textsc{IND}$) and specifically on the relations for the unseen labels ($\textsc{IND}\setminus\textsc{FIR}$). Given that all the previous models (DKRL, BLP, KEPLER, StAR, and StATIK) are inductive in entities but transductive in relations, none of them can make informed predictions in this setting, so we compare to a random baseline which computes the expected MRR for random rankings of candidate entities $\hat{\mathcal{E}}$ as follows:\nolinebreak
\vspace{-1ex}
\begin{equation*}
    \mathbb{E}[\text{MRR}_{\text{random}}] = \frac{1}{|\hat{\mathcal{E}}|} \sum_{i=1}^{|\hat{\mathcal{E}}|} \frac{1}{i}
\vspace{-1ex}
\end{equation*}
The results in Table \ref{tab:fully_inductive} show the performance of our model on this new setting. Our approach shows promising results as it outperforms the random baseline by a significant margin. However, the performance drops considerably when training on $\textsc{FIR}$ and evaluating on $\textsc{IND}\setminus\textsc{FIR}$, indicating that the model struggles with unseen relations. Notably, the results on the Wikidata-5M dataset are considerably better than those obtained on the WN18RR and FB15k-237 datasets, probably due to having relations with more descriptive texts.
Nevertheless, these results highlight the need for further research in developing models that can effectively generalize to unseen relations.

\section{Conclusion}
\label{sec:conclusion}

We presented a new Transformers-based approach to link prediction in text-attributed knowledge graphs that combines textual descriptions and graph structure in a fully inductive setting. Our Fast-and-Frugal Text-Graph (FnF-TG) Transformers outperform previous state-of-the-art models on three popular datasets, showcasing the importance of capturing rich structured information about entities and their relations. Our approach achieves superior performance while maintaining efficiency and scalability, making it a promising solution for large-scale knowledge graph applications. Moreover, our ablation studies provide insights into the key factors contributing to its effectiveness, demonstrating the value of each component in our model. Additionally, we proposed a new evaluation setting for fully inductive link prediction, where relations are also inductive, and demonstrated the potential of our approach in this setting. 

\section*{Limitations}
\label{sec:limitations}

While our approach has achieved promising results, there are opportunities for further improvement.

One area for exploration is optimising the scalability of our \textit{Graph Transformer Encoder} component (see Figure \ref{fig:fnf}), which currently requires computing fully quadratic attention over the entire ego-graph of a given entity. In fact it could still require considerable resources if the number of nodes in the ego-graph is scaled to the order of thousands, hundreds of thousands, or even millions.

Our work demonstrates that effectively capturing even local neighbourhood information is both non-trivial and under-explored and that it can significantly enhance performance. Indeed, our simplification to a 1-hop neighbourhood (ego-graph) was a careful decision to balance effectiveness and complexity. This approach not only allows for a fair comparison with the current state-of-the-art method, StATIK \citep{markowitz-etal-2022-statik}, but also mitigates the exponential increase in computational complexity (see Appendix Subsection \ref{subsec:complexity}) associated with larger neighbourhoods. While this predefined 1-hop neighbourhood provides a solid starting point, there is room to explore better alternatives. For instance, investigating multi-hop neighbourhoods or adaptive neighbourhood definitions could uncover more nuanced insights from the graph structure, potentially leading to even better results.

By building upon our framework, future work could refine these aspects, ultimately enhancing the effectiveness and versatility of our approach.

\section*{Ethics Statement}
\label{sec:ethics_statement}

We do not anticipate any ethical concerns related to our work, as it primarily presents an alternative approach to a previously proposed method. Our main contribution lies in introducing a new approach for link prediction. In our experiments, we use the same datasets and pretrained models as previous research, all of which are publicly available. However, it is important to acknowledge that these datasets and models may still require further examination for potential fairness issues and the knowledge they encapsulate.

\section*{Acknowledgements}
\label{sec:acknowledgements}
We extend our special gratitude to the Swiss National Science Foundation (SNSF) and Research Foundation – Flanders (FWO) for funding this work under grants 200021E\_189458 and G094020N.

\bibliography{anthology, custom}

\begin{thebibliography}{78}
\providecommand{\natexlab}[1]{#1}

\bibitem[{Balazevic et~al.(2018)Balazevic, Allen, and Hospedales}]{Balazevic2018HypernetworkKG}
Ivana Balazevic, Carl Allen, and Timothy~M. Hospedales. 2018.
\newblock \href {https://api.semanticscholar.org/CorpusID:52056218} {Hypernetwork knowledge graph embeddings}.
\newblock In \emph{International Conference on Artificial Neural Networks}.

\bibitem[{Banko et~al.(2007)Banko, Cafarella, Soderland, Broadhead, and Etzioni}]{Banko2007OpenIE}
Michele Banko, Michael~J. Cafarella, Stephen Soderland, Matthew Broadhead, and Oren Etzioni. 2007.
\newblock \href {https://api.semanticscholar.org/CorpusID:207169186} {Open information extraction from the web}.
\newblock In \emph{CACM}.

\bibitem[{Bhowmik and de~Melo(2020)}]{bhowmik2020explainable}
Rajarshi Bhowmik and Gerard de~Melo. 2020.
\newblock Explainable link prediction for emerging entities in knowledge graphs.
\newblock In \emph{The Semantic Web--ISWC 2020: 19th International Semantic Web Conference, Athens, Greece, November 2--6, 2020, Proceedings, Part I 19}, pages 39--55. Springer.

\bibitem[{Bordes et~al.(2013)Bordes, Usunier, Garcia-Duran, Weston, and Yakhnenko}]{NIPS2013_1cecc7a7}
Antoine Bordes, Nicolas Usunier, Alberto Garcia-Duran, Jason Weston, and Oksana Yakhnenko. 2013.
\newblock \href {https://proceedings.neurips.cc/paper_files/paper/2013/file/1cecc7a77928ca8133fa24680a88d2f9-Paper.pdf} {Translating embeddings for modeling multi-relational data}.
\newblock In \emph{Advances in Neural Information Processing Systems}, volume~26. Curran Associates, Inc.

\bibitem[{Bosselut et~al.(2019)Bosselut, Rashkin, Sap, Malaviya, Celikyilmaz, and Choi}]{bosselut-etal-2019-comet}
Antoine Bosselut, Hannah Rashkin, Maarten Sap, Chaitanya Malaviya, Asli Celikyilmaz, and Yejin Choi. 2019.
\newblock \href {https://doi.org/10.18653/v1/P19-1470} {{COMET}: Commonsense transformers for automatic knowledge graph construction}.
\newblock In \emph{Proceedings of the 57th Annual Meeting of the Association for Computational Linguistics}, pages 4762--4779, Florence, Italy. Association for Computational Linguistics.

\bibitem[{Brown et~al.(2020)Brown, Mann, Ryder, Subbiah, Kaplan, Dhariwal, Neelakantan, Shyam, Sastry, Askell, Agarwal, Herbert-Voss, Krueger, Henighan, Child, Ramesh, Ziegler, Wu, Winter, Hesse, Chen, Sigler, Litwin, Gray, Chess, Clark, Berner, McCandlish, Radford, Sutskever, and Amodei}]{NEURIPS2020_1457c0d6}
Tom Brown, Benjamin Mann, Nick Ryder, Melanie Subbiah, Jared~D Kaplan, Prafulla Dhariwal, Arvind Neelakantan, Pranav Shyam, Girish Sastry, Amanda Askell, Sandhini Agarwal, Ariel Herbert-Voss, Gretchen Krueger, Tom Henighan, Rewon Child, Aditya Ramesh, Daniel Ziegler, Jeffrey Wu, Clemens Winter, Chris Hesse, Mark Chen, Eric Sigler, Mateusz Litwin, Scott Gray, Benjamin Chess, Jack Clark, Christopher Berner, Sam McCandlish, Alec Radford, Ilya Sutskever, and Dario Amodei. 2020.
\newblock \href {https://proceedings.neurips.cc/paper_files/paper/2020/file/1457c0d6bfcb4967418bfb8ac142f64a-Paper.pdf} {Language models are few-shot learners}.
\newblock In \emph{Advances in Neural Information Processing Systems}, volume~33, pages 1877--1901. Curran Associates, Inc.

\bibitem[{Chen et~al.(2023)Chen, Cheng, Liu, Jiao, Ji, and Gao}]{Chen2023PretrainingTF}
Sanxing Chen, Hao Cheng, Xiaodong Liu, Jian Jiao, Yangfeng Ji, and Jianfeng Gao. 2023.
\newblock \href {https://api.semanticscholar.org/CorpusID:257771370} {Pre-training transformers for knowledge graph completion}.
\newblock \emph{ArXiv}, abs/2303.15682.

\bibitem[{Chen et~al.(2021)Chen, Liu, Gao, Jiao, Zhang, and Ji}]{chen-etal-2021-hitter}
Sanxing Chen, Xiaodong Liu, Jianfeng Gao, Jian Jiao, Ruofei Zhang, and Yangfeng Ji. 2021.
\newblock \href {https://doi.org/10.18653/v1/2021.emnlp-main.812} {{H}itt{ER}: Hierarchical transformers for knowledge graph embeddings}.
\newblock In \emph{Proceedings of the 2021 Conference on Empirical Methods in Natural Language Processing}, pages 10395--10407, Online and Punta Cana, Dominican Republic. Association for Computational Linguistics.

\bibitem[{Coman et~al.(2023)Coman, Barlacchi, and de~Gispert}]{coman-etal-2023-strong}
Andrei Coman, Gianni Barlacchi, and Adri{\`a} de~Gispert. 2023.
\newblock \href {https://doi.org/10.18653/v1/2023.findings-emnlp.417} {Strong and efficient baselines for open domain conversational question answering}.
\newblock In \emph{Findings of the Association for Computational Linguistics: EMNLP 2023}, pages 6305--6314, Singapore. Association for Computational Linguistics.

\bibitem[{Coman et~al.(2024)Coman, Theodoropoulos, Moens, and Henderson}]{coman-etal-2024-gadepo}
Andrei Coman, Christos Theodoropoulos, Marie-Francine Moens, and James Henderson. 2024.
\newblock \href {https://doi.org/10.18653/v1/2024.knowledgenlp-1.1} {{GAD}e{P}o: Graph-assisted declarative pooling transformers for document-level relation extraction}.
\newblock In \emph{Proceedings of the 3rd Workshop on Knowledge Augmented Methods for NLP}, pages 1--14, Bangkok, Thailand. Association for Computational Linguistics.

\bibitem[{Dai et~al.(2021)Dai, Zheng, Luo, Yang, Liu, Sui, and Chang}]{dai-etal-2021-inductively}
Damai Dai, Hua Zheng, Fuli Luo, Pengcheng Yang, Tianyu Liu, Zhifang Sui, and Baobao Chang. 2021.
\newblock \href {https://doi.org/10.18653/v1/2021.repl4nlp-1.10} {Inductively representing out-of-knowledge-graph entities by optimal estimation under translational assumptions}.
\newblock In \emph{Proceedings of the 6th Workshop on Representation Learning for NLP (RepL4NLP-2021)}, pages 83--89, Online. Association for Computational Linguistics.

\bibitem[{Dalton et~al.(2014)Dalton, Dietz, and Allan}]{10.1145/2600428.2609628}
Jeffrey Dalton, Laura Dietz, and James Allan. 2014.
\newblock \href {https://doi.org/10.1145/2600428.2609628} {Entity query feature expansion using knowledge base links}.
\newblock In \emph{Proceedings of the 37th International ACM SIGIR Conference on Research \& Development in Information Retrieval}, SIGIR '14, page 365–374, New York, NY, USA. Association for Computing Machinery.

\bibitem[{Daza et~al.(2021)Daza, Cochez, and Groth}]{10.1145/3442381.3450141}
Daniel Daza, Michael Cochez, and Paul Groth. 2021.
\newblock \href {https://doi.org/10.1145/3442381.3450141} {Inductive entity representations from text via link prediction}.
\newblock In \emph{Proceedings of the Web Conference 2021}, WWW '21, page 798–808, New York, NY, USA. Association for Computing Machinery.

\bibitem[{Dettmers et~al.(2018)Dettmers, Minervini, Stenetorp, and Riedel}]{10.5555/3504035.3504256}
Tim Dettmers, Pasquale Minervini, Pontus Stenetorp, and Sebastian Riedel. 2018.
\newblock Convolutional 2d knowledge graph embeddings.
\newblock In \emph{Proceedings of the Thirty-Second AAAI Conference on Artificial Intelligence and Thirtieth Innovative Applications of Artificial Intelligence Conference and Eighth AAAI Symposium on Educational Advances in Artificial Intelligence}, AAAI'18/IAAI'18/EAAI'18. AAAI Press.

\bibitem[{Devlin et~al.(2019)Devlin, Chang, Lee, and Toutanova}]{devlin-etal-2019-bert}
Jacob Devlin, Ming-Wei Chang, Kenton Lee, and Kristina Toutanova. 2019.
\newblock \href {https://doi.org/10.18653/v1/N19-1423} {{BERT}: Pre-training of deep bidirectional transformers for language understanding}.
\newblock In \emph{Proceedings of the 2019 Conference of the North {A}merican Chapter of the Association for Computational Linguistics: Human Language Technologies, Volume 1 (Long and Short Papers)}, pages 4171--4186, Minneapolis, Minnesota. Association for Computational Linguistics.

\bibitem[{Ebisu and Ichise(2018)}]{10.5555/3504035.3504257}
Takuma Ebisu and Ryutaro Ichise. 2018.
\newblock Toruse: knowledge graph embedding on a lie group.
\newblock In \emph{Proceedings of the Thirty-Second AAAI Conference on Artificial Intelligence and Thirtieth Innovative Applications of Artificial Intelligence Conference and Eighth AAAI Symposium on Educational Advances in Artificial Intelligence}, AAAI'18/IAAI'18/EAAI'18. AAAI Press.

\bibitem[{Elfwing et~al.(2017)Elfwing, Uchibe, and Doya}]{Elfwing2017SigmoidWeightedLU}
Stefan Elfwing, Eiji Uchibe, and Kenji Doya. 2017.
\newblock \href {https://api.semanticscholar.org/CorpusID:6940861} {Sigmoid-weighted linear units for neural network function approximation in reinforcement learning}.
\newblock \emph{Neural networks : the official journal of the International Neural Network Society}, 107:3--11.

\bibitem[{Falcon and {The PyTorch Lightning team}(2019)}]{Falcon_PyTorch_Lightning_2019}
William Falcon and {The PyTorch Lightning team}. 2019.
\newblock \href {https://doi.org/10.5281/zenodo.3828935} {{PyTorch Lightning}}.

\bibitem[{Fensel et~al.(2020)Fensel, Simsek, Angele, Huaman, K{\"a}rle, Panasiuk, Toma, Umbrich, and Wahler}]{Fensel2020KnowledgeGM}
Dieter~A. Fensel, Umutcan Simsek, Kevin Angele, Elwin Huaman, Elias K{\"a}rle, Oleksandra Panasiuk, Ioan Toma, J{\"u}rgen Umbrich, and Alexander Wahler. 2020.
\newblock \href {https://api.semanticscholar.org/CorpusID:210975360} {Knowledge graphs: Methodology, tools and selected use cases}.
\newblock \emph{Knowledge Graphs}.

\bibitem[{Galkin et~al.(2021)Galkin, Wu, Denis, and Hamilton}]{Galkin2021NodePieceCA}
Mikhail Galkin, Jiapeng Wu, E.~Denis, and William~L. Hamilton. 2021.
\newblock \href {https://api.semanticscholar.org/CorpusID:235606453} {Nodepiece: Compositional and parameter-efficient representations of large knowledge graphs}.
\newblock \emph{ArXiv}, abs/2106.12144.

\bibitem[{Gilmer et~al.(2017)Gilmer, Schoenholz, Riley, Vinyals, and Dahl}]{10.5555/3305381.3305512}
Justin Gilmer, Samuel~S. Schoenholz, Patrick~F. Riley, Oriol Vinyals, and George~E. Dahl. 2017.
\newblock Neural message passing for quantum chemistry.
\newblock In \emph{Proceedings of the 34th International Conference on Machine Learning - Volume 70}, ICML'17, page 1263–1272. JMLR.org.

\bibitem[{Gupta et~al.(2019)Gupta, Kenkre, and Talukdar}]{gupta-etal-2019-care}
Swapnil Gupta, Sreyash Kenkre, and Partha Talukdar. 2019.
\newblock \href {https://doi.org/10.18653/v1/D19-1036} {{C}a{R}e: Open knowledge graph embeddings}.
\newblock In \emph{Proceedings of the 2019 Conference on Empirical Methods in Natural Language Processing and the 9th International Joint Conference on Natural Language Processing (EMNLP-IJCNLP)}, pages 378--388, Hong Kong, China. Association for Computational Linguistics.

\bibitem[{Henderson et~al.(2023)Henderson, Mohammadshahi, Coman, and Miculicich}]{henderson-etal-2023-transformers}
James Henderson, Alireza Mohammadshahi, Andrei Coman, and Lesly Miculicich. 2023.
\newblock \href {https://doi.org/10.18653/v1/2023.bigpicture-1.8} {Transformers as graph-to-graph models}.
\newblock In \emph{Proceedings of the Big Picture Workshop}, pages 93--107, Singapore. Association for Computational Linguistics.

\bibitem[{Jiang et~al.(2022)Jiang, Zhou, Zhao, and Wen}]{jiang2022unikgqa}
Jinhao Jiang, Kun Zhou, Xin Zhao, and Ji-Rong Wen. 2022.
\newblock Unikgqa: Unified retrieval and reasoning for solving multi-hop question answering over knowledge graph.
\newblock In \emph{The Eleventh International Conference on Learning Representations}.

\bibitem[{Jiang et~al.(2019)Jiang, Wang, and Wang}]{jiang-etal-2019-adaptive}
Xiaotian Jiang, Quan Wang, and Bin Wang. 2019.
\newblock \href {https://doi.org/10.18653/v1/N19-1103} {Adaptive convolution for multi-relational learning}.
\newblock In \emph{Proceedings of the 2019 Conference of the North {A}merican Chapter of the Association for Computational Linguistics: Human Language Technologies, Volume 1 (Long and Short Papers)}, pages 978--987, Minneapolis, Minnesota. Association for Computational Linguistics.

\bibitem[{Kazemi and Poole(2018)}]{NEURIPS2018_b2ab0019}
Seyed~Mehran Kazemi and David Poole. 2018.
\newblock \href {https://proceedings.neurips.cc/paper_files/paper/2018/file/b2ab001909a8a6f04b51920306046ce5-Paper.pdf} {Simple embedding for link prediction in knowledge graphs}.
\newblock In \emph{Advances in Neural Information Processing Systems}, volume~31. Curran Associates, Inc.

\bibitem[{Ke et~al.(2021)Ke, Ji, Ran, Cui, Wang, Song, Zhu, and Huang}]{ke-etal-2021-jointgt}
Pei Ke, Haozhe Ji, Yu~Ran, Xin Cui, Liwei Wang, Linfeng Song, Xiaoyan Zhu, and Minlie Huang. 2021.
\newblock \href {https://doi.org/10.18653/v1/2021.findings-acl.223} {{J}oint{GT}: Graph-text joint representation learning for text generation from knowledge graphs}.
\newblock In \emph{Findings of the Association for Computational Linguistics: ACL-IJCNLP 2021}, pages 2526--2538, Online. Association for Computational Linguistics.

\bibitem[{Lin et~al.(2015)Lin, Liu, Sun, Liu, and Zhu}]{Lin_Liu_Sun_Liu_Zhu_2015}
Yankai Lin, Zhiyuan Liu, Maosong Sun, Yang Liu, and Xuan Zhu. 2015.
\newblock \href {https://doi.org/10.1609/aaai.v29i1.9491} {Learning entity and relation embeddings for knowledge graph completion}.
\newblock \emph{Proceedings of the AAAI Conference on Artificial Intelligence}, 29(1).

\bibitem[{Liu et~al.(2020)Liu, Jiang, He, Chen, Liu, Gao, and Han}]{Liu2020On}
Liyuan Liu, Haoming Jiang, Pengcheng He, Weizhu Chen, Xiaodong Liu, Jianfeng Gao, and Jiawei Han. 2020.
\newblock \href {https://openreview.net/forum?id=rkgz2aEKDr} {On the variance of the adaptive learning rate and beyond}.
\newblock In \emph{International Conference on Learning Representations}.

\bibitem[{Logan et~al.(2019)Logan, Liu, Peters, Gardner, and Singh}]{logan-etal-2019-baracks}
Robert Logan, Nelson~F. Liu, Matthew~E. Peters, Matt Gardner, and Sameer Singh. 2019.
\newblock \href {https://doi.org/10.18653/v1/P19-1598} {{B}arack{'}s wife hillary: Using knowledge graphs for fact-aware language modeling}.
\newblock In \emph{Proceedings of the 57th Annual Meeting of the Association for Computational Linguistics}, pages 5962--5971, Florence, Italy. Association for Computational Linguistics.

\bibitem[{Malaviya et~al.(2020)Malaviya, Bhagavatula, Bosselut, and Choi}]{malaviya2020commonsense}
Chaitanya Malaviya, Chandra Bhagavatula, Antoine Bosselut, and Yejin Choi. 2020.
\newblock Commonsense knowledge base completion with structural and semantic context.
\newblock \emph{Proceedings of the 34th AAAI Conference on Artificial Intelligence}.

\bibitem[{Markowitz et~al.(2022)Markowitz, Balasubramanian, Mirtaheri, Annavaram, Galstyan, and Ver~Steeg}]{markowitz-etal-2022-statik}
Elan Markowitz, Keshav Balasubramanian, Mehrnoosh Mirtaheri, Murali Annavaram, Aram Galstyan, and Greg Ver~Steeg. 2022.
\newblock \href {https://doi.org/10.18653/v1/2022.findings-naacl.46} {{S}t{ATIK}: Structure and text for inductive knowledge graph completion}.
\newblock In \emph{Findings of the Association for Computational Linguistics: NAACL 2022}, pages 604--615, Seattle, United States. Association for Computational Linguistics.

\bibitem[{Miculicich and Henderson(2022)}]{miculicich-henderson-2022-graph}
Lesly Miculicich and James Henderson. 2022.
\newblock \href {https://doi.org/10.18653/v1/2022.findings-acl.215} {Graph refinement for coreference resolution}.
\newblock In \emph{Findings of the Association for Computational Linguistics: ACL 2022}, pages 2732--2742, Dublin, Ireland. Association for Computational Linguistics.

\bibitem[{Mintz et~al.(2009)Mintz, Bills, Snow, and Jurafsky}]{mintz-etal-2009-distant}
Mike Mintz, Steven Bills, Rion Snow, and Daniel Jurafsky. 2009.
\newblock \href {https://aclanthology.org/P09-1113} {Distant supervision for relation extraction without labeled data}.
\newblock In \emph{Proceedings of the Joint Conference of the 47th Annual Meeting of the {ACL} and the 4th International Joint Conference on Natural Language Processing of the {AFNLP}}, pages 1003--1011, Suntec, Singapore. Association for Computational Linguistics.

\bibitem[{Mohammadshahi and Henderson(2020)}]{mohammadshahi-henderson-2020-graph}
Alireza Mohammadshahi and James Henderson. 2020.
\newblock \href {https://doi.org/10.18653/v1/2020.findings-emnlp.294} {Graph-to-graph transformer for transition-based dependency parsing}.
\newblock In \emph{Findings of the Association for Computational Linguistics: EMNLP 2020}, pages 3278--3289, Online. Association for Computational Linguistics.

\bibitem[{Mohammadshahi and Henderson(2021)}]{mohammadshahi-henderson-2021-recursive}
Alireza Mohammadshahi and James Henderson. 2021.
\newblock \href {https://doi.org/10.1162/tacl_a_00358} {Recursive non-autoregressive graph-to-graph transformer for dependency parsing with iterative refinement}.
\newblock \emph{Transactions of the Association for Computational Linguistics}, 9:120--138.

\bibitem[{Nickel et~al.(2015)Nickel, Murphy, Tresp, and Gabrilovich}]{nickel2015review}
Maximilian Nickel, Kevin Murphy, Volker Tresp, and Evgeniy Gabrilovich. 2015.
\newblock A review of relational machine learning for knowledge graphs.
\newblock \emph{Proceedings of the IEEE}, 104(1):11--33.

\bibitem[{Nickel et~al.(2011)Nickel, Tresp, and Kriegel}]{10.5555/3104482.3104584}
Maximilian Nickel, Volker Tresp, and Hans-Peter Kriegel. 2011.
\newblock A three-way model for collective learning on multi-relational data.
\newblock In \emph{Proceedings of the 28th International Conference on International Conference on Machine Learning}, ICML'11, page 809–816, Madison, WI, USA. Omnipress.

\bibitem[{Niu et~al.(2022)Niu, Li, Zhang, and Pu}]{niu-etal-2022-cake}
Guanglin Niu, Bo~Li, Yongfei Zhang, and Shiliang Pu. 2022.
\newblock \href {https://doi.org/10.18653/v1/2022.acl-long.205} {{CAKE}: A scalable commonsense-aware framework for multi-view knowledge graph completion}.
\newblock In \emph{Proceedings of the 60th Annual Meeting of the Association for Computational Linguistics (Volume 1: Long Papers)}, pages 2867--2877, Dublin, Ireland. Association for Computational Linguistics.

\bibitem[{Paszke et~al.(2019)Paszke, Gross, Massa, Lerer, Bradbury, Chanan, Killeen, Lin, Gimelshein, Antiga, Desmaison, Kopf, Yang, DeVito, Raison, Tejani, Chilamkurthy, Steiner, Fang, Bai, and Chintala}]{Paszke_PyTorch_An_Imperative_2019}
Adam Paszke, Sam Gross, Francisco Massa, Adam Lerer, James Bradbury, Gregory Chanan, Trevor Killeen, Zeming Lin, Natalia Gimelshein, Luca Antiga, Alban Desmaison, Andreas Kopf, Edward Yang, Zachary DeVito, Martin Raison, Alykhan Tejani, Sasank Chilamkurthy, Benoit Steiner, Lu~Fang, Junjie Bai, and Soumith Chintala. 2019.
\newblock \href {http://papers.neurips.cc/paper/9015-pytorch-an-imperative-style-high-performance-deep-learning-library.pdf} {{PyTorch: An Imperative Style, High-Performance Deep Learning Library}}.
\newblock In \emph{Advances in Neural Information Processing Systems 32}, pages 8024--8035. Curran Associates, Inc.

\bibitem[{Raffel et~al.(2019)Raffel, Shazeer, Roberts, Lee, Narang, Matena, Zhou, Li, and Liu}]{Raffel2019ExploringTL}
Colin Raffel, Noam~M. Shazeer, Adam Roberts, Katherine Lee, Sharan Narang, Michael Matena, Yanqi Zhou, Wei Li, and Peter~J. Liu. 2019.
\newblock \href {https://api.semanticscholar.org/CorpusID:204838007} {Exploring the limits of transfer learning with a unified text-to-text transformer}.
\newblock \emph{J. Mach. Learn. Res.}, 21:140:1--140:67.

\bibitem[{Saxena et~al.(2022)Saxena, Kochsiek, and Gemulla}]{saxena-etal-2022-sequence}
Apoorv Saxena, Adrian Kochsiek, and Rainer Gemulla. 2022.
\newblock \href {https://doi.org/10.18653/v1/2022.acl-long.201} {Sequence-to-sequence knowledge graph completion and question answering}.
\newblock In \emph{Proceedings of the 60th Annual Meeting of the Association for Computational Linguistics (Volume 1: Long Papers)}, pages 2814--2828, Dublin, Ireland. Association for Computational Linguistics.

\bibitem[{Schlichtkrull et~al.(2017)Schlichtkrull, Kipf, Bloem, van~den Berg, Titov, and Welling}]{Schlichtkrull2017ModelingRD}
M.~Schlichtkrull, Thomas Kipf, Peter Bloem, Rianne van~den Berg, Ivan Titov, and Max Welling. 2017.
\newblock \href {https://api.semanticscholar.org/CorpusID:5458500} {Modeling relational data with graph convolutional networks}.
\newblock In \emph{Extended Semantic Web Conference}.

\bibitem[{Schneider et~al.(2022)Schneider, Schopf, Vladika, Galkin, Simperl, and Matthes}]{schneider-etal-2022-decade}
Phillip Schneider, Tim Schopf, Juraj Vladika, Mikhail Galkin, Elena Simperl, and Florian Matthes. 2022.
\newblock \href {https://aclanthology.org/2022.aacl-main.46} {A decade of knowledge graphs in natural language processing: A survey}.
\newblock In \emph{Proceedings of the 2nd Conference of the Asia-Pacific Chapter of the Association for Computational Linguistics and the 12th International Joint Conference on Natural Language Processing (Volume 1: Long Papers)}, pages 601--614, Online only. Association for Computational Linguistics.

\bibitem[{Shah et~al.(2019)Shah, Villmow, Ulges, Schwanecke, and Shafait}]{shah2019open}
Haseeb Shah, Johannes Villmow, Adrian Ulges, Ulrich Schwanecke, and Faisal Shafait. 2019.
\newblock An open-world extension to knowledge graph completion models.
\newblock In \emph{Proceedings of the AAAI conference on artificial intelligence}, volume~33, pages 3044--3051.

\bibitem[{Shazeer(2020)}]{Shazeer2020GLUVI}
Noam~M. Shazeer. 2020.
\newblock \href {https://api.semanticscholar.org/CorpusID:211096588} {Glu variants improve transformer}.
\newblock \emph{ArXiv}, abs/2002.05202.

\bibitem[{Shi and Weninger(2018)}]{shi2018open}
Baoxu Shi and Tim Weninger. 2018.
\newblock Open-world knowledge graph completion.
\newblock In \emph{Proceedings of the AAAI conference on artificial intelligence}, volume~32.

\bibitem[{Socher et~al.(2013)Socher, Chen, Manning, and Ng}]{NIPS2013_b337e84d}
Richard Socher, Danqi Chen, Christopher~D Manning, and Andrew Ng. 2013.
\newblock \href {https://proceedings.neurips.cc/paper_files/paper/2013/file/b337e84de8752b27eda3a12363109e80-Paper.pdf} {Reasoning with neural tensor networks for knowledge base completion}.
\newblock In \emph{Advances in Neural Information Processing Systems}, volume~26. Curran Associates, Inc.

\bibitem[{Su et~al.(2021)Su, Lu, Pan, Wen, and Liu}]{Su2021RoFormerET}
Jianlin Su, Yu~Lu, Shengfeng Pan, Bo~Wen, and Yunfeng Liu. 2021.
\newblock \href {https://api.semanticscholar.org/CorpusID:233307138} {Roformer: Enhanced transformer with rotary position embedding}.
\newblock \emph{ArXiv}, abs/2104.09864.

\bibitem[{Sun et~al.(2019)Sun, Deng, Nie, and Tang}]{sun2019rotate}
Zhiqing Sun, Zhi-Hong Deng, Jian-Yun Nie, and Jian Tang. 2019.
\newblock Rotate: Knowledge graph embedding by relational rotation in complex space.
\newblock \emph{arXiv preprint arXiv:1902.10197}.

\bibitem[{Teru et~al.(2020)Teru, Denis, and Hamilton}]{10.5555/3524938.3525814}
Komal~K. Teru, Etienne~G. Denis, and William~L. Hamilton. 2020.
\newblock Inductive relation prediction by subgraph reasoning.
\newblock In \emph{Proceedings of the 37th International Conference on Machine Learning}, ICML'20. JMLR.org.

\bibitem[{Theodoropoulos et~al.(2021)Theodoropoulos, Henderson, Coman, and Moens}]{theodoropoulos-etal-2021-imposing}
Christos Theodoropoulos, James Henderson, Andrei~Catalin Coman, and Marie-Francine Moens. 2021.
\newblock \href {https://doi.org/10.18653/v1/2021.conll-1.27} {Imposing relation structure in language-model embeddings using contrastive learning}.
\newblock In \emph{Proceedings of the 25th Conference on Computational Natural Language Learning}, pages 337--348, Online. Association for Computational Linguistics.

\bibitem[{Toutanova and Chen(2015)}]{toutanova-chen-2015-observed}
Kristina Toutanova and Danqi Chen. 2015.
\newblock \href {https://doi.org/10.18653/v1/W15-4007} {Observed versus latent features for knowledge base and text inference}.
\newblock In \emph{Proceedings of the 3rd Workshop on Continuous Vector Space Models and their Compositionality}, pages 57--66, Beijing, China. Association for Computational Linguistics.

\bibitem[{Touvron et~al.(2023)Touvron, Lavril, Izacard, Martinet, Lachaux, Lacroix, Rozi{\`e}re, Goyal, Hambro, Azhar, Rodriguez, Joulin, Grave, and Lample}]{Touvron2023LLaMAOA}
Hugo Touvron, Thibaut Lavril, Gautier Izacard, Xavier Martinet, Marie-Anne Lachaux, Timoth{\'e}e Lacroix, Baptiste Rozi{\`e}re, Naman Goyal, Eric Hambro, Faisal Azhar, Aurelien Rodriguez, Armand Joulin, Edouard Grave, and Guillaume Lample. 2023.
\newblock \href {https://api.semanticscholar.org/CorpusID:257219404} {Llama: Open and efficient foundation language models}.
\newblock \emph{ArXiv}, abs/2302.13971.

\bibitem[{Trouillon et~al.(2016)Trouillon, Welbl, Riedel, Gaussier, and Bouchard}]{trouillon2016complex}
Th{\'e}o Trouillon, Johannes Welbl, Sebastian Riedel, {\'E}ric Gaussier, and Guillaume Bouchard. 2016.
\newblock Complex embeddings for simple link prediction.
\newblock In \emph{International conference on machine learning}, pages 2071--2080. PMLR.

\bibitem[{Turc et~al.(2019)Turc, Chang, Lee, and Toutanova}]{Turc2019WellReadSL}
Iulia Turc, Ming-Wei Chang, Kenton Lee, and Kristina Toutanova. 2019.
\newblock \href {https://api.semanticscholar.org/CorpusID:202889175} {Well-read students learn better: On the importance of pre-training compact models}.
\newblock \emph{arXiv: Computation and Language}.

\bibitem[{Vaswani et~al.(2017)Vaswani, Shazeer, Parmar, Uszkoreit, Jones, Gomez, Kaiser, and Polosukhin}]{NIPS2017_3f5ee243}
Ashish Vaswani, Noam Shazeer, Niki Parmar, Jakob Uszkoreit, Llion Jones, Aidan~N Gomez, \L~ukasz Kaiser, and Illia Polosukhin. 2017.
\newblock \href {https://proceedings.neurips.cc/paper_files/paper/2017/file/3f5ee243547dee91fbd053c1c4a845aa-Paper.pdf} {Attention is all you need}.
\newblock In \emph{Advances in Neural Information Processing Systems}, volume~30. Curran Associates, Inc.

\bibitem[{Wang et~al.(2021{\natexlab{a}})Wang, Shen, Long, Zhou, Wang, and Chang}]{10.1145/3442381.3450043}
Bo~Wang, Tao Shen, Guodong Long, Tianyi Zhou, Ying Wang, and Yi~Chang. 2021{\natexlab{a}}.
\newblock \href {https://doi.org/10.1145/3442381.3450043} {Structure-augmented text representation learning for efficient knowledge graph completion}.
\newblock In \emph{Proceedings of the Web Conference 2021}, WWW '21, page 1737–1748, New York, NY, USA. Association for Computing Machinery.

\bibitem[{Wang et~al.(2020)Wang, Ren, and Leskovec}]{Wang2020RelationalMP}
Hongwei Wang, Hongyu Ren, and Jure Leskovec. 2020.
\newblock \href {https://api.semanticscholar.org/CorpusID:235262529} {Relational message passing for knowledge graph completion}.
\newblock \emph{Proceedings of the 27th ACM SIGKDD Conference on Knowledge Discovery \& Data Mining}.

\bibitem[{Wang et~al.(2022)Wang, Zhao, Wei, and Liu}]{wang-etal-2022-simkgc}
Liang Wang, Wei Zhao, Zhuoyu Wei, and Jingming Liu. 2022.
\newblock \href {https://doi.org/10.18653/v1/2022.acl-long.295} {{S}im{KGC}: Simple contrastive knowledge graph completion with pre-trained language models}.
\newblock In \emph{Proceedings of the 60th Annual Meeting of the Association for Computational Linguistics (Volume 1: Long Papers)}, pages 4281--4294, Dublin, Ireland. Association for Computational Linguistics.

\bibitem[{Wang et~al.(2019{\natexlab{a}})Wang, Han, Li, and Pan}]{10.1609/aaai.v33i01.33017152}
Peifeng Wang, Jialong Han, Chenliang Li, and Rong Pan. 2019{\natexlab{a}}.
\newblock \href {https://doi.org/10.1609/aaai.v33i01.33017152} {Logic attention based neighborhood aggregation for inductive knowledge graph embedding}.
\newblock In \emph{Proceedings of the Thirty-Third AAAI Conference on Artificial Intelligence and Thirty-First Innovative Applications of Artificial Intelligence Conference and Ninth AAAI Symposium on Educational Advances in Artificial Intelligence}, AAAI'19/IAAI'19/EAAI'19. AAAI Press.

\bibitem[{Wang et~al.(2019{\natexlab{b}})Wang, Huang, Wang, Dai, Jiang, Liu, Lyu, Zhu, and Wu}]{wang2019coke}
Quan Wang, Pingping Huang, Haifeng Wang, Songtai Dai, Wenbin Jiang, Jing Liu, Yajuan Lyu, Yong Zhu, and Hua Wu. 2019{\natexlab{b}}.
\newblock Coke: Contextualized knowledge graph embedding.
\newblock \emph{arXiv preprint arXiv:1911.02168}.

\bibitem[{Wang et~al.(2017)Wang, Mao, Wang, and Guo}]{8047276}
Quan Wang, Zhendong Mao, Bin Wang, and Li~Guo. 2017.
\newblock \href {https://doi.org/10.1109/TKDE.2017.2754499} {Knowledge graph embedding: A survey of approaches and applications}.
\newblock \emph{IEEE Transactions on Knowledge and Data Engineering}, 29(12):2724--2743.

\bibitem[{Wang et~al.(2021{\natexlab{b}})Wang, Gao, Zhu, Zhang, Liu, Li, and Tang}]{wang-etal-2021-kepler}
Xiaozhi Wang, Tianyu Gao, Zhaocheng Zhu, Zhengyan Zhang, Zhiyuan Liu, Juanzi Li, and Jian Tang. 2021{\natexlab{b}}.
\newblock \href {https://doi.org/10.1162/tacl_a_00360} {{KEPLER}: A unified model for knowledge embedding and pre-trained language representation}.
\newblock \emph{Transactions of the Association for Computational Linguistics}, 9:176--194.

\bibitem[{Wolf et~al.(2020)Wolf, Debut, Sanh, Chaumond, Delangue, Moi, Cistac, Rault, Louf, Funtowicz, Davison, Shleifer, von Platen, Ma, Jernite, Plu, Xu, Le~Scao, Gugger, Drame, Lhoest, and Rush}]{wolf-etal-2020-transformers}
Thomas Wolf, Lysandre Debut, Victor Sanh, Julien Chaumond, Clement Delangue, Anthony Moi, Pierric Cistac, Tim Rault, Remi Louf, Morgan Funtowicz, Joe Davison, Sam Shleifer, Patrick von Platen, Clara Ma, Yacine Jernite, Julien Plu, Canwen Xu, Teven Le~Scao, Sylvain Gugger, Mariama Drame, Quentin Lhoest, and Alexander Rush. 2020.
\newblock \href {https://doi.org/10.18653/v1/2020.emnlp-demos.6} {Transformers: State-of-the-art natural language processing}.
\newblock In \emph{Proceedings of the 2020 Conference on Empirical Methods in Natural Language Processing: System Demonstrations}, pages 38--45, Online. Association for Computational Linguistics.

\bibitem[{Xie et~al.(2016)Xie, Liu, Jia, Luan, and Sun}]{Xie2016RepresentationLO}
Ruobing Xie, Zhiyuan Liu, Jia Jia, Huanbo Luan, and Maosong Sun. 2016.
\newblock \href {https://api.semanticscholar.org/CorpusID:31606602} {Representation learning of knowledge graphs with entity descriptions}.
\newblock In \emph{AAAI Conference on Artificial Intelligence}.

\bibitem[{Xiong et~al.(2020)Xiong, Yang, He, Zheng, Zheng, Xing, Zhang, Lan, Wang, and Liu}]{Xiong2020OnLN}
Ruibin Xiong, Yunchang Yang, Di~He, Kai Zheng, Shuxin Zheng, Chen Xing, Huishuai Zhang, Yanyan Lan, Liwei Wang, and Tie-Yan Liu. 2020.
\newblock \href {https://api.semanticscholar.org/CorpusID:211082816} {On layer normalization in the transformer architecture}.
\newblock \emph{ArXiv}, abs/2002.04745.

\bibitem[{Xu et~al.(2021)Xu, Wang, Lyu, Zhu, and Mao}]{Xu2021EntitySW}
Benfeng Xu, Quan Wang, Yajuan Lyu, Yong Zhu, and Zhendong Mao. 2021.
\newblock \href {https://api.semanticscholar.org/CorpusID:231985811} {Entity structure within and throughout: Modeling mention dependencies for document-level relation extraction}.
\newblock In \emph{AAAI Conference on Artificial Intelligence}.

\bibitem[{Yang et~al.(2014)Yang, Yih, He, Gao, and Deng}]{yang2014embedding}
Bishan Yang, Wen-tau Yih, Xiaodong He, Jianfeng Gao, and Li~Deng. 2014.
\newblock Embedding entities and relations for learning and inference in knowledge bases.
\newblock \emph{arXiv preprint arXiv:1412.6575}.

\bibitem[{Yang et~al.(2024)Yang, Liu, Zhang, Zhang, Xie, and Mao}]{yang-etal-2024-knowledge}
Guangqian Yang, Yi~Liu, Lei Zhang, Licheng Zhang, Hongtao Xie, and Zhendong Mao. 2024.
\newblock \href {https://doi.org/10.18653/v1/2024.findings-acl.509} {Knowledge context modeling with pre-trained language models for contrastive knowledge graph completion}.
\newblock In \emph{Findings of the Association for Computational Linguistics: ACL 2024}, pages 8619--8630, Bangkok, Thailand. Association for Computational Linguistics.

\bibitem[{Yang et~al.(2022)Yang, Gupta, Upadhyay, He, Goel, and Paul}]{yang-etal-2022-tableformer}
Jingfeng Yang, Aditya Gupta, Shyam Upadhyay, Luheng He, Rahul Goel, and Shachi Paul. 2022.
\newblock \href {https://doi.org/10.18653/v1/2022.acl-long.40} {{T}able{F}ormer: Robust transformer modeling for table-text encoding}.
\newblock In \emph{Proceedings of the 60th Annual Meeting of the Association for Computational Linguistics (Volume 1: Long Papers)}, pages 528--537, Dublin, Ireland. Association for Computational Linguistics.

\bibitem[{Yang et~al.(2023)Yang, Chen, Li, Ding, and Wu}]{Yang2023GiveUT}
Lin~F. Yang, Hongyang Chen, Zhao Li, Xiao Ding, and Xindong Wu. 2023.
\newblock \href {https://api.semanticscholar.org/CorpusID:259203671} {Give us the facts: Enhancing large language models with knowledge graphs for fact-aware language modeling}.
\newblock \emph{IEEE Transactions on Knowledge and Data Engineering}, 36:3091--3110.

\bibitem[{Yao et~al.(2019)Yao, Mao, and Luo}]{Yao2019KGBERTBF}
Liang Yao, Chengsheng Mao, and Yuan Luo. 2019.
\newblock \href {https://api.semanticscholar.org/CorpusID:202539519} {Kg-bert: Bert for knowledge graph completion}.
\newblock \emph{ArXiv}, abs/1909.03193.

\bibitem[{Ying et~al.(2021)Ying, Cai, Luo, Zheng, Ke, He, Shen, and Liu}]{Ying2021DoTR}
Chengxuan Ying, Tianle Cai, Shengjie Luo, Shuxin Zheng, Guolin Ke, Di~He, Yanming Shen, and Tie-Yan Liu. 2021.
\newblock \href {https://api.semanticscholar.org/CorpusID:235376852} {Do transformers really perform bad for graph representation?}
\newblock In \emph{Neural Information Processing Systems}.

\bibitem[{Yu et~al.(2022)Yu, Zhu, Fang, Yu, Wang, Xu, Ren, Yang, and Zeng}]{yu-etal-2022-kg}
Donghan Yu, Chenguang Zhu, Yuwei Fang, Wenhao Yu, Shuohang Wang, Yichong Xu, Xiang Ren, Yiming Yang, and Michael Zeng. 2022.
\newblock \href {https://doi.org/10.18653/v1/2022.acl-long.340} {{KG}-{F}i{D}: Infusing knowledge graph in fusion-in-decoder for open-domain question answering}.
\newblock In \emph{Proceedings of the 60th Annual Meeting of the Association for Computational Linguistics (Volume 1: Long Papers)}, pages 4961--4974, Dublin, Ireland. Association for Computational Linguistics.

\bibitem[{Zha et~al.(2021)Zha, Chen, and Yan}]{Zha2021InductiveRP}
Hanwen Zha, Zhiyu Chen, and Xifeng Yan. 2021.
\newblock \href {https://api.semanticscholar.org/CorpusID:232222958} {Inductive relation prediction by bert}.
\newblock \emph{ArXiv}, abs/2103.07102.

\bibitem[{Zhang et~al.(2020)Zhang, Liu, Xiong, and Liu}]{zhang-etal-2020-grounded}
Houyu Zhang, Zhenghao Liu, Chenyan Xiong, and Zhiyuan Liu. 2020.
\newblock \href {https://doi.org/10.18653/v1/2020.acl-main.184} {Grounded conversation generation as guided traverses in commonsense knowledge graphs}.
\newblock In \emph{Proceedings of the 58th Annual Meeting of the Association for Computational Linguistics}, pages 2031--2043, Online. Association for Computational Linguistics.

\bibitem[{Zhu et~al.(2021)Zhu, Zhang, Xhonneux, and Tang}]{Zhu2021NeuralBN}
Zhaocheng Zhu, Zuobai Zhang, Louis-Pascal Xhonneux, and Jian Tang. 2021.
\newblock \href {https://api.semanticscholar.org/CorpusID:235422273} {Neural bellman-ford networks: A general graph neural network framework for link prediction}.
\newblock In \emph{Neural Information Processing Systems}.

\end{thebibliography}

\appendix

\section{Appendix}
\label{sec:appendix}

\subsection{Datasets statistics}

Table \ref{tab:datasets} provides the statistics of the datasets used in our experiments. $\mathcal{E}$ represents the set of entities, $\mathcal{R}$ denotes the set of relation labels, $\mathcal{T}$ consists of the set of relation triples $(h,r,t) \in \mathcal{E} \times \mathcal{R} \times \mathcal{E}$, and $E(e)$ shows the mean and standard deviation ($\mu_{\sigma}$) of the number of neighbours in an entity's ego-graph.

\begin{table*}[tb]
  \centering
  \scalebox{0.75}{
      \begin{tabular}{l|c|ccc|ccc|ccc}
        \hline
        \textbf{Dataset} & $\mathcal{R}$ & $\mathcal{E}_{train}$ & $\mathcal{T}_{train}$ & $E(e)_{train}$ & $\mathcal{E}_{val}$ & $\mathcal{T}_{val}$ & $E(e)_{val}$ & $\mathcal{E}_{test}$ & $\mathcal{T}_{test}$ & $E(e)_{test}$ \\
        \hline
        WN18RR\textsubscript{\textsc{IND}} & $11$ & $32,755$ & $69,585$ & $2,12_{3,15}$ & $4,094$ & $11,381$ & $1,17_{1,33}$ & $4,456$ & $12,037$ & $1,18_{1,35}$ \\
        FB15K-237\textsubscript{\textsc{IND}} & $237$ & $11,633$ & $215,082$ & $18,49_{28,91}$ & $1,454$ & $42,164$ & $4,70_{10,63}$ & $2,416$ & $52,870$ & $4,97_{12,29}$ \\
        Wikidata-5M\textsubscript{\textsc{IND}} & $822$ & $4,579,609$ & $20,496,514$ & $4,48_{4,41}$ & $7,374$ & $6,699$ & $0,91_{0,78}$ & $7,475$ & $6,894$ & $0,92_{0,81}$ \\
        \hline
      \end{tabular}
  }
  \caption{
    WN18RR\textsubscript{\textsc{IND}}, FB15K-237\textsubscript{\textsc{IND}}, and Wikidata-5M\textsubscript{\textsc{IND}} datasets statistics. $\mathcal{E}$ represents the set of entities, $\mathcal{R}$ denotes the set of relation labels, $\mathcal{T}$ consists of the set of relation triples $(h,r,t) \in \mathcal{E} \times \mathcal{R} \times \mathcal{E}$, and $E(e)$ shows the mean and standard deviation ($\mu_{\sigma}$) of the number of neighbours in an entity's ego-graph.
  }
  \label{tab:datasets}
\end{table*}

\subsection{Complexity of FnF-TG}
\label{subsec:complexity}

Our method exhibits identical computational complexity to StATIk \citep{markowitz-etal-2022-statik}, with $O(N + Q)$ complexity, where $N$ denotes the number of nodes in the graph and $Q$ represents the number of queries $(h, r, \hat{e})$ and $(\hat{e}, r, t)$.

\subsection{Training and Implementation Details}

We provide details on the training and implementation of our models on three datasets: WN18RR\textsubscript{\textsc{IND}}, FB15k-237\textsubscript{\textsc{IND}}, and Wikidata5M\textsubscript{\textsc{IND}}.

\paragraph{Seeds and Epochs} We run our experiments with five different seeds ($73, 21, 37, 3, 7$) for WN18RR\textsubscript{\textsc{IND}} and FB15k-237\textsubscript{\textsc{IND}}, and two seeds ($73, 21$) for Wikidata5M\textsubscript{\textsc{IND}} due to its large scale (see Table \ref{tab:datasets}). We train our models for 40 epochs on WN18RR\textsubscript{\textsc{IND}} and FB15k-237\textsubscript{\textsc{IND}}, and 5 epochs on Wikidata5M\textsubscript{\textsc{IND}}, following previous works \citep{10.1145/3442381.3450141, markowitz-etal-2022-statik}.

\paragraph{Hyperparameters} We set the number of sampled neighbors per entity based on the dataset statistics (Table \ref{tab:datasets}): $10$ for WN18RR\textsubscript{\textsc{IND}}, $40$ for FB15k-237\textsubscript{\textsc{IND}}, and $1$ for Wikidata5M\textsubscript{\textsc{IND}}. We use $24$ words of text for each $x_{KG}$ in WN18RR\textsubscript{\textsc{IND}} and FB15k-237\textsubscript{\textsc{IND}}, and $64$ words for Wikidata5M\textsubscript{\textsc{IND}}.

\paragraph{Graph Transformer Encoder} We implement the \textit{Graph Transformer Encoder} layer using a pre-LayerNorm Transformer \citep{Xiong2020OnLN} with a SwiGLU-type pointwise feed-forward network \cite{Shazeer2020GLUVI}. We use a single GT layer, as multiple layers did not improve performance while increasing latency.

\paragraph{Optimisation} We set the learning rate to $1e^{-5}$ for a batch size of $32$ and scale it proportionally with the batch size following a power-of-2 rule to fit the GPU budget. We use RAdam \cite{Liu2020On} as our optimiser and a cosine learning rate decay throughout the training process.

\paragraph{Libraries} We develop our models using PyTorch \cite{Paszke_PyTorch_An_Imperative_2019}, Lightning \cite{Falcon_PyTorch_Lightning_2019}, and Hugging Face's Transformers \cite{wolf-etal-2020-transformers} libraries.

\subsection{Computational Budget}

We fix our computational budget to a constant consumer-grade GPU (NVIDIA RTX3090 24GB) as stated in Subsection \ref{subsec:controlled_experimental_setup} and report the GPU budget per run for each dataset on $\text{FnF-TG}_{\text{BERT}\textsubscript{\textsc{base}}}$ relative to the largest text encoders. The GPU budget per run is 4 GPU/h for WN18RR\textsubscript{\textsc{IND}}, 6 GPU/h for FB15k-237\textsubscript{\textsc{IND}}, and 40 GPU/h for Wikidata5M\textsubscript{\textsc{IND}}.

\end{document}